\documentclass[twoside,11pt]{article}

\usepackage{jmlr2e}

\usepackage{graphicx}
\usepackage{caption}
\usepackage{subcaption}
\usepackage{blindtext}
\usepackage{amsmath}
\usepackage{multirow}

\makeatletter
\renewcommand*{\thanks}[1]{%
  \footnotemark
  \protected@xdef\@thanks{\@thanks
    \protect\footnotetext[\arabic{footnote}]{#1}}%
}
\makeatother

\usepackage[svgnames]{xcolor}
\usepackage[framemethod=tikz]{mdframed}
\usepackage[customcolors]{hf-tikz}
\usetikzlibrary{shadows}

\newcounter{exa}

\newmdenv[
settings={\refstepcounter{exa}},
linewidth=1pt,
innertopmargin=1.5\baselineskip,
roundcorner=6pt,
backgroundcolor=yellow!10,
linecolor=red!70!black,
frametitle=example~\theexa,
frametitlefont=\scshape\color{red!70!black}
]{mdexample}

\usepackage{lastpage}

\firstpageno{1}

\begin{document}

\title{ABHINAW \thanks{\textbf{\textit{A}}utomated \textbf{\textit{B}}enchmarking like \textbf{\textit{H}}umans for \textbf{I}mage \textbf{\textit{N}}otion within \textbf{\textit{A}}I-generated \textbf{\textit{W}}ords/Typography} : A method for Automatic Evaluation of Typography within AI-Generated Images}

\author{\name \name Abhinaw Jagtap\email 2022pai9032@iitjammu.ac.in \\
       \addr Department of Computer Science and Engineering\\
       Indian Institute of Technology\\
       Jammu, INDIA
       \AND
       \name Nachiket Tapas
       \email nachikettapas.cse@csvtu.ac.in \\
       \addr  Department of Computer Science and Engineering \\
       Chhattisgarh Swami Vivekanand Technical University (CSVTU)\\
       Bhilai, India
       \AND
       \name R.G. Brajesh
       \email brajeshrg@csvtu.ac.in \\
       \addr  Department of Bio Medical Engineering \\
       Chhattisgarh Swami Vivekanand Technical University (CSVTU)\\
       Bhilai, India
       }
  
      \editor{NA}

\maketitle

\begin{abstract}
In the fast-evolving field of Generative AI, platforms like MidJourney, DALL-E, and Stable Diffusion have transformed Text-to-Image (T2I) Generation. 
However, despite their impressive ability to create high-quality images, they often struggle to generate accurate text within these images. 
Theoretically, if we could achieve accurate text generation in AI images in a ``zero-shot'' manner, it would not only make AI-generated images more meaningful but also democratize the graphic design industry. 
The first step towards this goal is to create a robust scoring matrix for evaluating text accuracy in AI-generated images. 
Although there are existing bench-marking methods like CLIP SCORE and T2I-CompBench++, there's still a gap in systematically evaluating text and typography in AI-generated images, especially with diffusion-based methods.
In this paper, we introduce a novel evaluation matrix designed explicitly for quantifying the performance of text and typography generation within AI-generated images. 
We have used letter by letter matching strategy to compute the exact matching scores from the reference text to the AI generated text. 
Our novel approach to calculate the score takes care of multiple redundancies such as repetition of words, case sensitivity, mixing of words, irregular incorporation of letters etc. 
Moreover, we have developed a Novel method named as brevity adjustment to handle excess text.
In addition we have also done a quantitative analysis of frequent errors arise due to frequently used words and less frequently used words. 
Project page is available at: \url{https://github.com/Abhinaw3906/ABHINAW-MATRIX}
\end{abstract}


\section{Introduction}
In present world almost every computation task can be automated even if the person who is operating is with very low skill due to the limitless potential of AI. 
Similarly, the prospect of achieving flawless textual integration within images generated by diffusion models sparks considerable excitement, heralding a transformative moment for the graphic design industry. 
Imagine a scenario, where AI systems could replace the need for extensive design platforms, enabling users to generate complex graphics from a simple text prompt in a ``zero-shot'' fashion. 
This potential leap forward could democratize design, making it as accessible as typing a sentence, provided that the text within these AI-generated images meets high standards of accuracy and aesthetic integration.

\begin{figure}[h]
\centering
\includegraphics[width=0.8\textwidth]{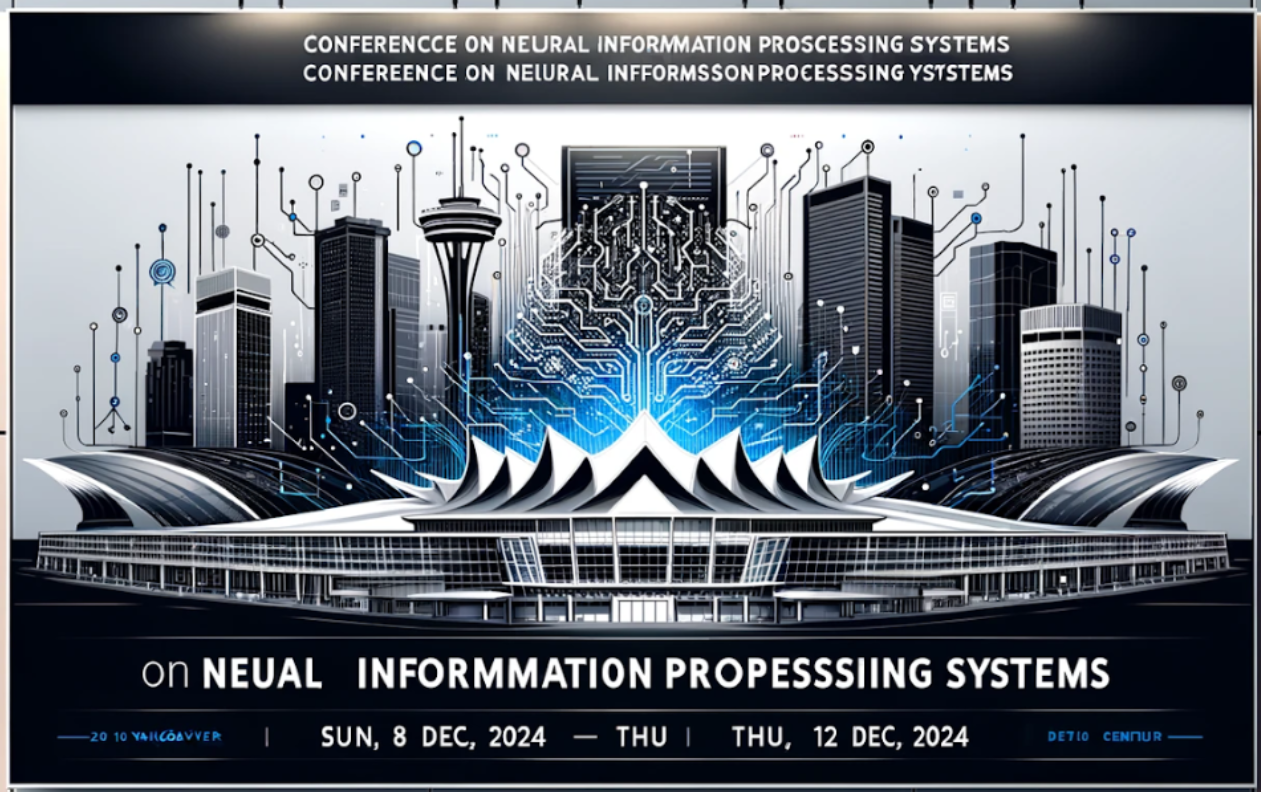}
\caption{Observe the generated image and consider the impact if every word were spelled correctly. The potential for such technology to craft fully realized graphic designs from simple prompts could significantly shift the current landscape of graphic design tools, providing a powerful, intuitive alternative to traditional methods.}
\label{fig:dalle_infographic}
\end{figure}

As we can see in Figure~\ref{fig:dalle_infographic} that presents an output example from DALL-E 3 for the prompt: ``Create an image infographic banner with text: `Neural Information Processing Systems (NeurIPS): Conference on Neural Information Processing Systems, Venue: Vancouver Convention Center, Date: Sun, 8 Dec, 2024 – Thu, 12 Dec, 2024'.'' 

\subsection{Rationale}
\begin{figure}[h]
\centering
\includegraphics[width=0.8\textwidth]{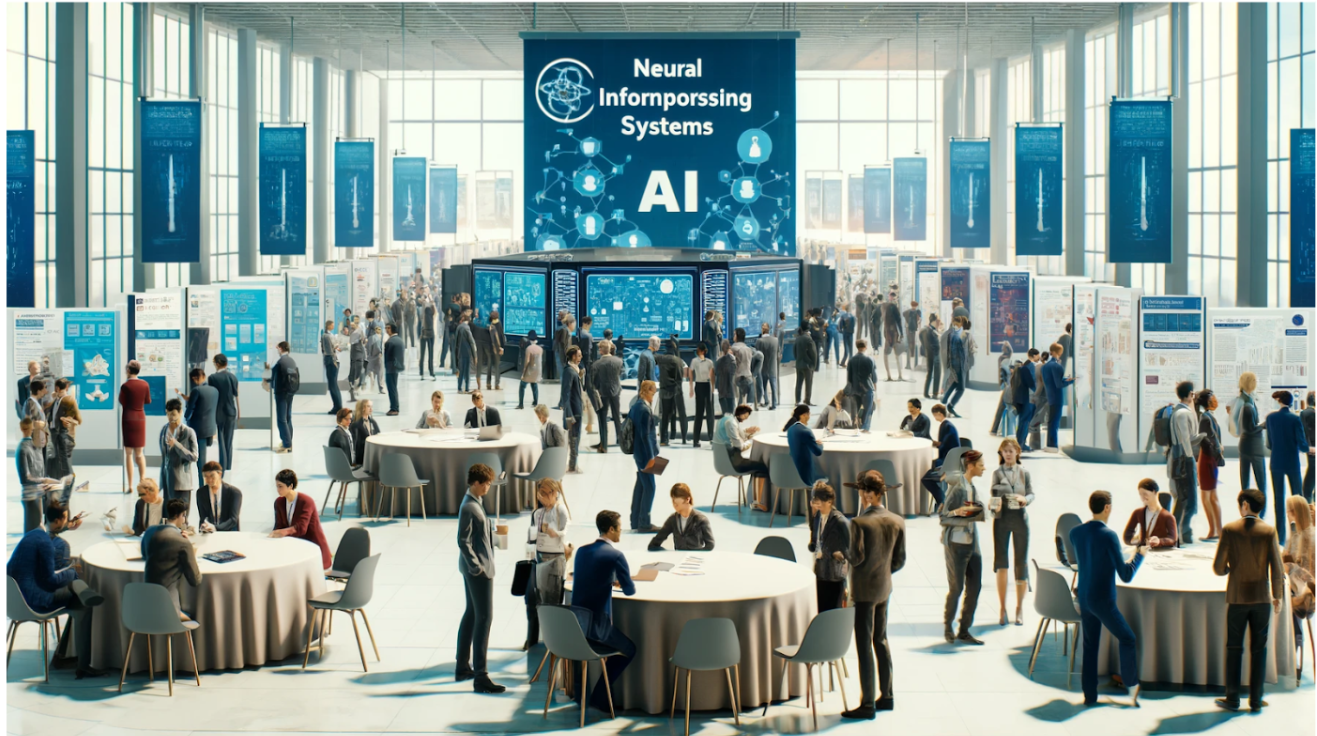}
\caption{\textit{\textbf{Dall-E 3} - instead of the text  \textbf{``Neural Information Processing Systems (NeurIPS)''} it has written the text as \textbf{``Nural Inforporcing Systems AI''}}}
\label{fig:dalle}
\end{figure}

The integration of text within visually generated content by AI has seen significant advancements with the advent of diffusion models. 
These models, which generate structured data from noise by reversing the process from noise to structured data, have enabled the creation of high-resolution images and videos from textual inputs. As demonstrated in Figure \ref{fig:dalle}, we utilized DALL-E 3's text-to-image capabilities to generate the visual content.
Recent works by ~\cite{saharia2022imagen} and ~\cite{ramesh2022hierarchical}, and others have established diffusion models as the de-facto standard for such tasks, showcasing their ability to produce images that are not only visually appealing but also contextually coherent with the input prompts.

However, despite these advancements, there exists a notable gap in the systematic evaluation of how text and typography are synthesized within these images. 
Historical benchmarks in related fields, such as the BLEU score for natural language processing, have catalyzed significant progress by providing clear, quantifiable metrics that align closely with human judgment~\cite{papineni-etal-2002-bleu}. 
In contrast, the field of text-to-image (T2I) generation currently lacks a robust, automated metric to evaluate the fidelity and legibility of text rendered within generated images. 
This gap not only impedes the ability to benchmark and compare different models effectively but also slows the iterative improvement of these systems.

Likewise, metrics like the Structural Similarity Index (SSIM) and Peak Signal-to-Noise Ratio (PSNR) have significantly influenced the assessment of image and video quality \cite{zhou2004image, huynh2008scope}. 
However, these metrics do not fully address the unique challenges of evaluating text in T2I outputs, such as font accuracy, text sharpness, and overall correctness of text, because even a single spelling mistake will make the output useless.
The limitations of these traditional metrics underscore the necessity for a specialized evaluation framework designed explicitly for text within AI-generated images. 
Addressing this gap is essential for advancing T2I technology by providing a clear standard for model comparison and improvement.

The current landscape of T2I (Text-to-Image) generation tools includes a variety of open-source and commercial platforms such as \textbf{MidJourney}~(\cite{midjourney}), \textbf{DALL-E}~(\cite{ramesh2022dalle2}), \textbf{Stable Diffusion}~(\cite{rombach2022high}), and others, which differ in their handling of text within images. 
For instance, preliminary comparisons, such as Figure \ref{fig:dalle} showing the output of DALL-E 3 for the \textbf{prompt: \textit{Generate image of conference with text “Neural Information Processing Systems (NeurIPS)”},} alongside similar outputs from other platforms, reveal varying degrees of accuracy and aesthetic integration of text. 
This inconsistency underscores the need for a standardized evaluation matrix.

\section{Related work}
\textbf{Text-to-Image generation:}
Early research in this domain, as evidenced by studies~(\cite{reed2016generative,reed2016learning,zhang2017stackgan,xu2018attngan,zhu2019dm,zhang2021cross}), explored various network architectures and loss functions leveraging generative adversarial networks (GANs)~(\cite{goodfellow2014generative}).
DALL$\cdot$E~(\cite{ramesh2021zero}) achieves impressive results with a transformer and discrete variational autoencoder (VAE) trained on web-scale data.
Recently, diffusion models have demonstrated significant advancements in text-to-image (T2I) generation~(\cite{ramesh2022hierarchical,nichol2022glide,rombach2022high,saharia2022imagen,gafni2022make}).
Models such as Stable Diffusion~(\cite{rombach2022high, podell2023sdxl, esser2024scaling}), DALL-E-3~(\cite{betker2023improving}), MDM~(\cite{gu2023matryoshka}), and Pixart-$\alpha$~(\cite{chen2023pixart}), trained on large-scale web data, have exhibited impressive generative capabilities.
Despite these advancements, state-of-the-art models like Stable Diffusion~(\cite{rombach2022high}) still face challenges in generating accurate text within AI Generated images.
Some recent efforts aim to align T2I models with human feedback~(\cite{zhang2023hive,lee2023aligning}). 
For instance, RAFT~(\cite{dong2023raft}) introduces reward-ranked fine-tuning to align these models with specific metrics.

\textbf{Compositional text-to-image generation.}
Researchers have investigated various aspects of compositionality to achieve visually coherent and semantically consistent results in T2I generation~(\cite{liu2022compositional,feng2022training,li2022stylet2i,wu2023harnessing}). 
Previous studies focused on concept conjunction and negation~(\cite{liu2022compositional}), attribute binding with colors~(\cite{feng2022training,chefer2023attend,park2021benchmark}), generative numeracy~(\cite{lian2023llm}), and spatial relationships between objects~(\cite{chen2023trainingfree,wu2023harnessing}).
However, these studies each targeted specific sub-problems and conducted evaluations in constrained scenarios. 
Recent compositional studies generally fall into two categories~(\cite{wang2024compositional}): those that rely on cross-attention maps for compositional generation~(\cite{meral2024conform, kim2023dense, rassin2024linguistic}), and those that integrate layout as a generation condition~(\cite{gani2023llm, li2023gligen, taghipour2024box, wang2024divide, chen2023reason}). 
Techniques like LMD~(\cite{lian2023llm}), RPG~(\cite{yang2024mastering}), and RealCompo~(\cite{zhang2024realcompo}) utilize LLMs or MLLMs to reason out layouts or decompose compositional problems.

\textbf{Benchmarks for text-to-image generation.}
Initial benchmarks for T2I models were evaluated on datasets like CUB birds~(\cite{wah2011caltech}), Oxford flowers~(\cite{nilsback2008automated}), and COCO~(\cite{lin2014microsoft}), which offer limited diversity. 
As T2I models have improved, more challenging benchmarks have been developed. 
DrawBench~(\cite{saharia2022imagen}) includes 200 prompts to evaluate various skills such as counting, compositions, conflicting scenarios, and writing. 
DALL-EVAL~(\cite{cho2022dall}) proposes PaintSkills to assess visual reasoning, image-text alignment, image quality, and social bias using 7,330 prompts.
HE-T2I~(\cite{petsiuk2022human}) suggests 900 prompts to evaluate counting, shapes, and faces for T2I models. 
Several compositional benchmarks have also emerged.
Park et al.~(\cite{park2021benchmark}) proposed benchmarks on CUB Birds~(\cite{wah2011caltech}) and Oxford Flowers~(\cite{nilsback2008automated}) to evaluate the models' ability to generate images with object-color and object-shape compositions. 
ABC-6K and CC500~(\cite{feng2022training}) benchmarks assess attribute binding but focus only on color attributes. 
HRS-Bench~(\cite{bakr2023hrs}) is a general-purpose benchmark evaluating 13 skills with 45,000 prompts.

\textbf{Evaluation metrics for text-to-image generation.}
Current evaluation metrics for T2I generation are categorized into fidelity assessment, alignment assessment, and LLM-based metrics. 
Traditional metrics such as Inception Score (IS)~(\cite{salimans2016improved}) and Frechet Inception Distance (FID)~(\cite{heusel2018gans}) are commonly used to evaluate the fidelity of synthesized images.
For assessing image-text alignment, techniques such as CLIP~(\cite{radford2021learning}), BLIP2~(\cite{li2023blip}) and text-text similarity using BLIP~(\cite{li2022blip}) captioning and CLIP text similarity are prevalent.
Some studies leverage the reasoning abilities of large language models (LLMs) for evaluation~(\cite{lu2023llmscore,chen2023xiqe}).
Additionally, human preferences or feedback are often included in the evaluation process~(\cite{xu2024imagereward, sun2023dreamsync, kirstain2023pick, wu2023better, wu2023human, liang2024rich, ku2023imagenhub, ku2023viescore}). 
More granular metrics have been proposed in~(\cite{lee2024holistic}). 
However, a study on the efficacy of text in these evaluation metrics for T2I generation is lacking. 
That is why we propose evaluation metrics specifically designed for text within AI generated images.

\section{Problem overview}
For any evaluation matrix to be efficacious, it is imperative to delineate a precise and unambiguous reference point for comparative analysis. 
In the domain of AI-generated text within images, establishing this baseline is crucial for ensuring that evaluations are both accurate and reproducible. 
To this end, for the nascent deployment of the ABHINAW Matrix, we utilize a simple yet potent mechanism to define the expected textual content within the generated images. 
\textbf{Our model employs a keyword, ``text," followed by the desired text encapsulated within double quotes.} 
This formatting signals to the AI that the enclosed text is the exact content required to appear in the generated image, thereby facilitating~\textbf{ease of automation} and providing a clear directive for the evaluation algorithms.
\\
For example, if the prompt is:
\\
\textbf{Prompt 1 : Create an image of people enjoying with text “Celebrate Freedom”}
\\
\textbf{Prompt 2 :} \textbf{Create a banner about text “Sale ends Sunday!”}

The AI is expected to generate an image wherein the phrase~\textbf{`Celebrate Freedom'} is accurately depicted within the visual context.  
In another prompt, the evaluation matrix would check for the precise appearance of text  \textbf{`Sale ends Sunday!'} in the output image. In the subsequent sections of this paper, this will be referred to as the ``Reference Text". 
Having clarified the concept of the Reference Text, we will now proceed to explore examples of traditional scoring matrices, followed by a discussion of their limitations. 

The Simple precision comparison method aims to  evaluate the typographical accuracy by aligning each character in the generated text with its counterpart in a predetermined reference text.
Under this conventional approach, this method would entail a character-by-character comparison, as shown in Table~\ref{tab:prec_comp}, where each correct match scores a point and mismatches do not. 
For instance, the alignment of 'S' in both the reference and the generated text would accrue one point. 
Conversely, the second character 'w' in the generated text diverges from the 'w' in the reference, resulting in zero points. The precision for the compared text is mentioned in the Table~\ref{tab:prec_calc}

\begin{table}[h]
\centering
\begin{tabular}{|c|c|c|c|c|c|c|c|}
\hline
Serial No. & 1 & 2 & 3 & 4 & 5 & 6 & 7 \\
\hline
Reference text & G & a & m & e & & o & n \\
\hline
Generated text & G & a & m & a & & o & n\\
\hline
Points & 1 & 1 & 1& 0 & 1& 1& 1\\
\hline
\end{tabular}
\caption{Comparison of reference and generated text}
\label{tab:prec_comp}
\end{table}

\begin{table}[h]
\centering
\begin{tabular}{|c|c|c|} \hline 
\textbf{Correct Text} & \textbf{Total Text in Reference} & \textbf{Precision in percentage} \\ \hline 
 6 & 7 & 6 / 7 * 100 = 85.71\% \\ 
\hline 
\end{tabular}
\caption{Precision calculation}
\label{tab:prec_calc}
\end{table}
To illustrate the application of simple precision, one might calculate the ratio of matching characters to the total number of characters in the reference text. as shown in Table \ref{tab:prec_calc}. 
This yields a numerical precision score for that particular image generation instance. 
However, this method reveals significant limitations. 
These limitations are discussed below:

\subsection{Case-sensitive precision inadequacy \label{subsec:Case-Sensitive precision inadequacy}}

Evaluating text within AI-generated images requires more than just traditional precision metrics; it necessitates a consideration of case insensitivity to maintain the creative liberty of AI image generators. 
The following points examine the importance of making the evaluation case-insensitive and how this adjustment affects its accuracy. 

\textit{Creative Liberty: }AI image generators often employ creative typography that includes varying cases for stylistic effects. 
Evaluating such text with case-sensitive precision can unfairly penalize the output.
For example, the reference text ``Neural Information Processing Systems" and the generated text ``neural information processing systems" are both identical in content.
However, a case-sensitive comparison would mark them as different, reducing the precision score.        
\textit{Consistency in Evaluation: }Case-insensitive evaluation ensures that variations in capitalization do not affect the accuracy score, leading to more consistent and fair assessments.
This approach aligns better with human evaluation, where the focus is on content rather than case.

\subsection {Lack of consideration for text arrangement\label{subsec:Lack of Consideration for Text Arrangement}}
Normal precision fails to recognize the creative arrangement of text in AI-generated images, which can mislead traditional OCR systems or even modern day Vision transformers, reducing the apparent precision, even if they are depicting the required text.

\begin{figure}[h]
    \centering
    \begin{subfigure}[b]{0.45\textwidth}
        \centering
        \includegraphics[width=\textwidth]{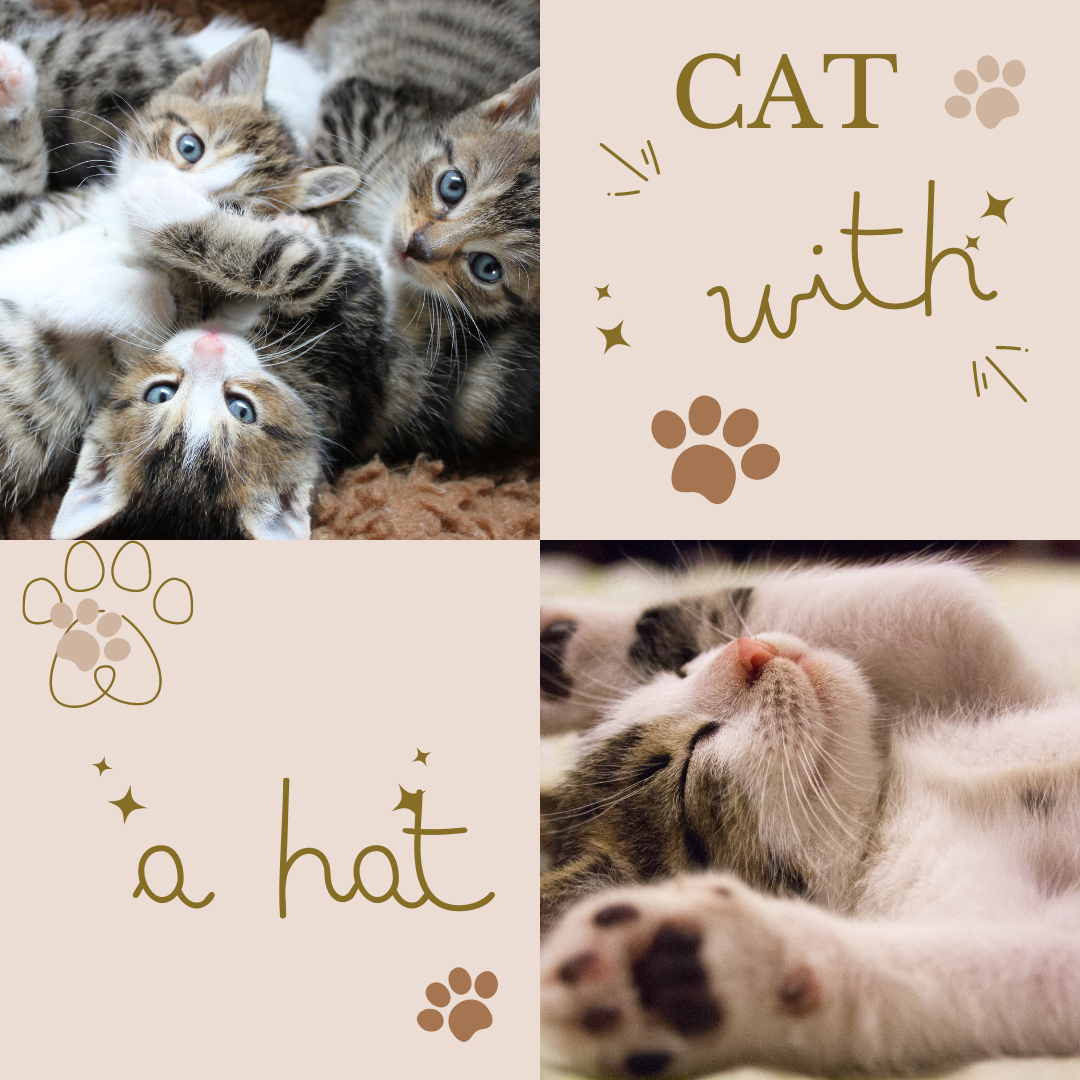}
        \caption{Output 1}
        \label{fig:ref_image}
    \end{subfigure}
    \hfill
    \begin{subfigure}[b]{0.45\textwidth}
        \centering
        \includegraphics[width=\textwidth]{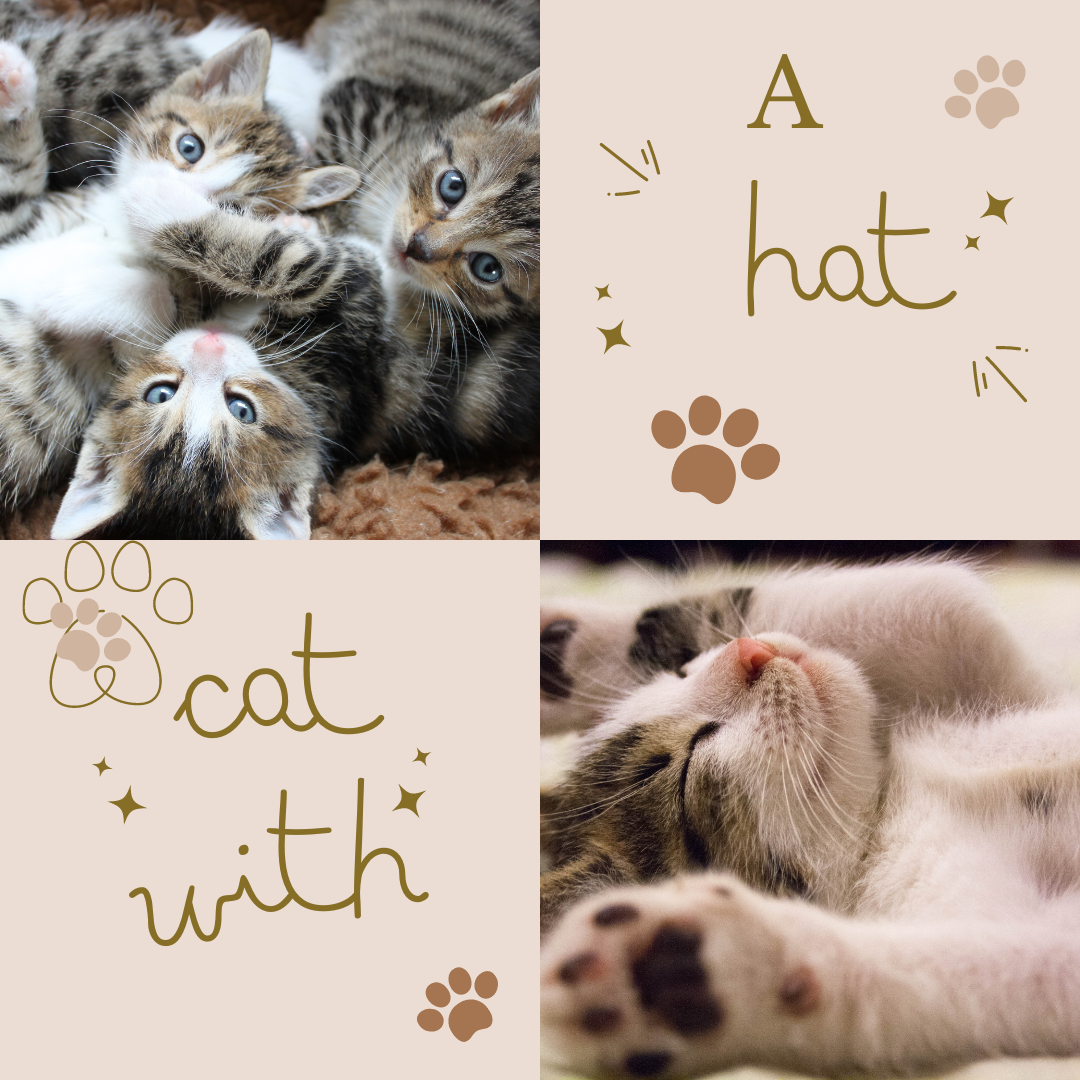}
        \caption{Output 2}
        \label{fig:gen_image}
    \end{subfigure}
    \caption{Side-by-side comparison of 2 images with similar and accurate texts.}
    \label{fig:comparison}
\end{figure}

\begin{figure}[h]
\centering
\includegraphics[width=0.8\textwidth]{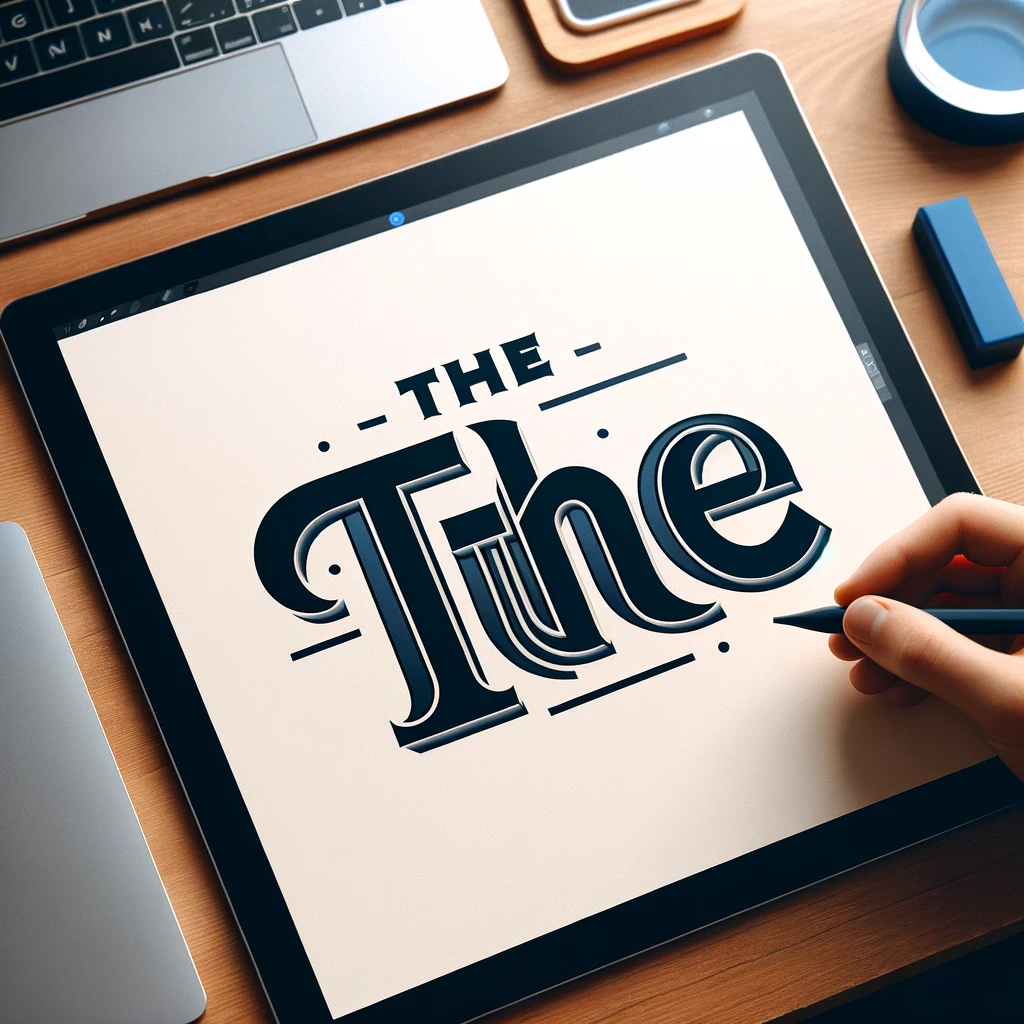}
\caption{\textit{Given the reference string ``the," the AI-generated output by DALL-E 3 resulted in the candidate string ``the the."}}
\label{fig:the}
\end{figure}

For instance, consider Output 1: ``CAT WITH A HAT" and Output 2: ``A HAT CAT WITH"  (Refer Figure ~\ref{fig:comparison}).
In the conventional approach, the method involves a character-by-character comparison, awarding points for each correct match. While this method may yield a precision score of 100\% for Output 1, it results in a significantly lower precision score for Output 2. This discrepancy occurs despite the fact that Output 2 also accurately conveys the intended text, with the difference being merely an issue of alignment. If the alignment were corrected, a typical human reader would interpret Output 2 as ``cat with a hat," demonstrating that it too meets the desired output for AI-generated images. 

\subsection{Inadequacy in addressing error significance due to excess or insufficient text}
Traditional precision metrics fail to consider the contextual significance of errors, particularly when AI erroneously generates more or less text than the Reference Text. This oversight can result in misleading evaluations.For example, When the candidate text is shorter than the reference text, traditional precision metrics address the issue by introducing extra spaces in the candidate text for comparison. This approach naturally reduces the precision score since the generated text is compared against blank spaces, which penalizes the absence of content. This method works well in scenarios where the candidate text is shorter, as it captures the gap between the reference and generated texts, leading to a more accurate evaluation.
However, a more complex problem arises when the candidate text is longer than the reference text. In such cases, traditional precision metrics may fail to properly penalize the extra content. Since the comparison is based on the length of the reference text, only the initial segment of the candidate text is evaluated. If this segment matches the reference text, it may result in a precision score of 100\%, as mentioned in the Table~\ref{tab:the_the_comparison} , which is misleading. This score does not account for the additional, erroneous content in the candidate text, thus failing to accurately reflect the significance of the error introduced
For example, reference text: ``the" and generated text: ``the the"  (Refer Figure ~\ref{fig:the}).the evaluation would consider the first ``the" as a perfect match, ignoring the additional ``the," leading to a misleadingly high precision score.

\noindent \textbf{Sum of Matches}: \( \sum_{i=1}^{n} \delta(R[i], C[i]) = 3 \) 
\\
Here:

\begin{itemize}
    \item \( R[i] \) refers to the \( i \)-th character of the reference text.
    
    \item \( C[i] \) refers to the \( i \)-th character of the generated text.
    
    \item \( \delta(R[i], C[i]) \) is a function defined as:
    \[
    \delta(R[i], C[i]) =
    \begin{cases} 
      1 & \text{if } R[i] = C[i], \\
      0 & \text{otherwise}.
    \end{cases}
    \]
    It returns 1 if the characters at position \( i \) in both the reference and the generated text match, and 0 otherwise.
    
    \item The sum \( \sum_{i=1}^{n} \delta(R[i], C[i]) \) is taken over all characters in the reference text, where \( n \) is the length of the reference text.
    
    \item When the candidate text is shorter than the reference text, traditional precision metrics address the issue by introducing extra spaces in the candidate text for comparison. Mathematically, this can be represented as:
    \[
    C'[i] =
    \begin{cases} 
      C[i] & \text{if } i \leq \text{length}(C), \\
      \text{(space)} & \text{if } i > \text{length}(C).
    \end{cases}
    \]
    
\end{itemize}

\begin{table}[h]
\centering
\begin{tabular}{|c|c|c|c|c|c|c|c|}
\hline
\textbf{Position (i)} & 1 & 2 & 3 & 4 & 5 & 6 & 7 \\
\hline
\textbf{Reference Text} & t & h & e &  &  &  &  \\
\hline
\textbf{Generated Text} & t & h & e & \text{(space)} & t & h & e \\
\hline
\end{tabular}
\caption{Comparison of reference and generated text for precision calculation}
\label{tab:the_the_comparison}
\end{table}

\noindent \textbf{Precision}: \( \frac{3}{3} = 1 \)\\

While the precision score is perfect, it doesn't reflect the contextual error introduced by the extra characters ``the"

\subsection{Inconsistency across multiple generations}
AI-generated images often exhibit variability with each generation, leading to fluctuating precision scores.
For example, the reference text: ``Neural Information Processing Systems," the generated text (1st iteration): ``Nural Inforporcing Systems," and the generated text (2nd iteration): ``Neeral infconriatitema syesy3ans" results in different scores each time of generation.

\section{Handling the problems associated with the simple precision}
In this section, we will discuss the methods to handle issues in the traditional scoring matrix and create a quantitative framework for the ABHINAW matrix.

\subsection{Handling case sensitivity}
To implement case-insensitive precision, we convert both the reference and candidate texts to a uniform case (either lower or upper) before performing the comparison. 
This ensures that the evaluation focuses on the textual content rather than its presentation.\\
Let \( R \) be the reference text and \( C \) be the candidate text. 
We convert both texts to lowercase (or uppercase) for comparison.
This adjustment ensures that the stylistic use of case does not affect the evaluation score.

\subsection{Addressing text arrangement in AI-generated images using cosine similarity}
In the evaluation of text within AI-generated images, normal precision and case-insensitive precision metrics can sometimes fall short, particularly when the arrangement of text is creatively modified. 
This section introduces the use of cosine similarity to address this limitation, ensuring a more robust evaluation metric.

AI-generated images often display text in creative and unconventional arrangements, which can lead to discrepancies in OCR (Optical Character Recognition) outputs. 
For instance, in the example have we have already seen earlier we consider the prompt "Generate an image of a cat with text 'cat with a hat'" and the AI-generated text "cat a hat with" arranged creatively within the image. 
Traditional precision metrics would mark this as incorrect despite the presence of all required words.

To address the issue of text arrangement, we use cosine similarity. 
Cosine similarity measures the cosine of the angle between two vectors in a multidimensional space, effectively capturing the similarity between two texts irrespective of their order.\\

\noindent \textbf{Mathematical formulation}:\\

\noindent Let \( R \) be the reference text and \( C \) be the candidate text. We represent both texts as term frequency vectors \( \mathbf{v_R} \) and \( \mathbf{v_C} \), respectively.

The first step would be converting the texts to vector

\[
\mathbf{v_R} = [f_1(R), f_2(R), \ldots, f_n(R)]
\]
\[
\mathbf{v_C} = [f_1(C), f_2(C), \ldots, f_n(C)]
\]

where \( f_i(T) \) represents the frequency of term \( i \) in text \( T \).

In the second step we will calculate the cosine similarity using following mathematical formulations 

\[
\text{cosine\_similarity} = \frac{\mathbf{v_R} \cdot \mathbf{v_C}}{||\mathbf{v_R}|| \times ||\mathbf{v_C}||}
\]

where \( \mathbf{v_R} \cdot \mathbf{v_C} \) is the dot product of the vectors, and \( ||\mathbf{v_R}|| \) and \( ||\mathbf{v_C}|| \) are the magnitudes of the vectors.\\

To integrate cosine similarity with normal precision, we use a conditional approach. If the cosine similarity is greater than a specified threshold (e.g., 0.9), we use the cosine similarity score. Otherwise, we fall back to the traditional precision metric.\\

The conditional approch can be easily presented with mathematical framework given below:

Let \( P \) be the precision score and \( CS \) be the cosine similarity. The final score \( S \) is given by:

\[
S = 
\begin{cases} 
CS & \text{if } CS > 0.9 \\
P & \text{if } CS \leq 0.9
\end{cases}
\]
We choose the threshold of 0.9 for cosine similarity because it indicates a very high contextual similarity between the reference and candidate texts. When the cosine similarity is above 0.9, we can be confident that the candidate text closely matches the context of the reference text, even if the word order is different. If the cosine similarity is below this threshold, it suggests that the candidate text deviates significantly from the reference, and normal precision should be used to ensure accuracy.

Let's understand this with an example:

\begin{itemize}
    \item \textbf{Reference Text}: "cat with a hat"
    \item \textbf{Generated Text}: "cat a hat with"
\end{itemize}

Based on the steps provided earlier, lets convert the text into vectors and then calculate the cosine similarity

\begin{enumerate}
    \item \textbf{Convert Texts to Vectors:}
        \begin{itemize}
            \item Term frequency vectors for "cat with a hat":
        \end{itemize}

        \[
        \mathbf{v_R} = [1, 1, 1, 1] \quad (\text{cat, with, a, hat})
        \]

        \begin{itemize}
            \item Term frequency vectors for "cat a hat with":
        \end{itemize}

        \[
        \mathbf{v_C} = [1, 1, 1, 1] \quad (\text{cat, a, hat, with})
        \]

    \item \textbf{Calculate cosine similarity:}

        \[
        \text{cosine\_similarity} = \frac{[1, 1, 1, 1] \cdot [1, 1, 1, 1]}{\sqrt{1^2 + 1^2 + 1^2 + 1^2} \times \sqrt{1^2 + 1^2 + 1^2 + 1^2}} = \frac{4}{\sqrt{4} \times \sqrt{4}} = \frac{4}{4} = 1
        \]

        Since the cosine similarity is 1, it is greater than 0.9.

    \item \textbf{Calculate precision (for comparison):}

        \[
        \text{Precision} = \frac{\text{Correct Characters}}{\text{Total Characters in Reference}} = \frac{4}{14} \approx 0.28
        \]

    \item \textbf{Apply conditional scoring:}

        \[
        S = 
        \begin{cases} 
        1 & \text{if } CS > 0.9 \\
        0.28 & \text{if } CS \leq 0.9
        \end{cases}
        \]

        Since the cosine similarity is 1, the final score \( S \) is 1.
\end{enumerate}

\textbf{Comparison without cosine similarity}

Without using cosine similarity, we would rely solely on precision for the evaluation. In the case of "cat with a hat" and "cat a hat with", the normal precision would be calculated as:

\[
\text{Precision} = \frac{\text{Correct Characters}}{\text{Total Characters in Reference}} = \frac{3}{14} \approx 0.21
\]

This score would unfairly penalize the text arrangement, despite the content being correct and contextually similar.

\begin{itemize}
    \item \textbf{Reference Text}: "cat with a hat"
    \item \textbf{Generated Text}: "cat a hat with"
\end{itemize}

Normal precision would yield:

\[
\text{Precision} = \frac{3}{14} \approx 0.21
\]

This result shows that without cosine similarity, we would not accurately capture the contextual similarity between the texts, leading to a lower precision score even though the image generation is contextually accurate.

\subsection{Addressing redundancy in AI-generated text:incorporating brevity adjustment}
\label{subsec:brevity_adjustment}

In the previous sections, we discussed the limitations of normal precision and the benefits of incorporating cosine similarity for handling creative text arrangements. However, cosine similarity alone cannot differentiate between correct and redundant text. This section introduces the brevity adjustment factor to address this issue, ensuring a more comprehensive evaluation metric.\\

While cosine similarity effectively captures contextual similarity, it fails to penalize redundant text. For example, consider the reference text "the" and the candidate text "the the". Both texts would have a cosine similarity of 1, indicating perfect similarity, which is misleading as the candidate text contains unnecessary repetition.\\

In traditional evaluation metrics like BLEU, the brevity penalty is used to address cases where candidate translations are excessively long compared to their references.\cite{papineni-etal-2002-bleu}\\

Our novel Brevity Adjustment (BA) method addresses a key limitation in traditional metrics like BLEU's brevity penalty (BP). BP is designed to penalize candidate texts that are shorter than their references, addressing over-generation errors. However, BA tackles the opposite problem—cases where the reference text is shorter than the candidate text. For example, when the reference is "the" and the candidate is "the the," BP would not penalize the redundancy, while our BA method would correctly identify and adjust for the error, ensuring a more accurate evaluation of translation quality.\\

\textbf{Mathematical Formulation}

Let \( n \) be the length of the reference text and \( m \) be the length of the candidate text. The brevity adjustment is defined as:

\[
\text{BA} = 
\begin{cases} 
1 & \text{if } m < n \\
e^{1 - \frac{m}{n}} & \text{if } m \geq n 
\end{cases}
\]

where \( e \) is the base of the natural logarithm.

The final score \( S \) is adjusted by multiplying the cosine similarity or precision by the brevity adjustment factor.

Let \( P \) be the precision score and \( CS \) be the cosine similarity. The final score \( S \) is given by:

\[
S = 
\begin{cases} 
CS \times \text{BA} & \text{if } CS > 0.9 \\
P \times \text{BA} & \text{if } CS \leq 0.9 
\end{cases}
\]

\textbf{Scenario 1}
\begin{itemize}
    \item \textbf{Reference Text}: ``the"
    \item \textbf{Candidate Text}: ``the the"
\end{itemize}

\textbf{Calculations:}

\begin{enumerate}
    \item Convert Texts to Vectors:
        \begin{itemize}
            \item Term frequency vectors for "the":
        \end{itemize}

        \[
        \mathbf{v_R} = [1] \quad (\text{the})
        \]
        
        \begin{itemize}
            \item Term frequency vectors for "the the":
        \end{itemize}
        
        \[
        \mathbf{v_C} = [2] \quad (\text{the})
        \]

    \item Calculate Cosine Similarity:
        \[
        \text{cosine\_similarity} = \frac{[1] \cdot [2]}{\sqrt{1^2} \sqrt{2^2}} = \frac{2}{1 \cdot 2} = \frac{2}{2} = 1
        \]

Since the cosine similarity is 1, it is greater than 0.9.
 
    \item Calculate Brevity Adjustment:
        \[
        n = 3, \quad m = 7
        \]
        \[
        \text{BA} = e^{1 - \frac{7}{3}} = e^{-1.33} \approx 0.2644
        \]

    \item Calculate Final Score:
            \[
            S = CS \times \text{BA} = 1 \times 0.2644 = 0.2644
            \]
            
This result correctly penalizes the candidate text for its redundancy.

\end{enumerate}

\textbf{Scenario 2}\\

Consider a reference text "the" and multiple candidate strings:
\begin{itemize}
    \item \textbf{Reference Text}: "the"
    \item \textbf{Candidate Strings}: "the the", "the"
\end{itemize}

\textbf{Calculations:}

\begin{enumerate}
    \item \textbf{Candidate 1}: "the the"
    \[
    \text{cosine\_similarity} = 1, \quad \text{BA} = e^{1 - \frac{7}{3}} = e^{-1.33} \approx 0.2644
    \]
    \[
    S = 1 \times 0.2644 \approx 0.2644
    \]

    \item \textbf{Candidate 2}: "the"
    \[
    \text{cosine\_similarity} = 1, \quad \text{BA} = 1
    \]
    \[
    S = 1 \times 1 = 1
    \]

\end{enumerate}

Incorporating brevity adjustment into the evaluation of AI-generated text effectively addresses the issue of redundancy, complementing the use of cosine similarity for handling creative text arrangements. This approach ensures a more comprehensive and accurate assessment of text fidelity, aligning closely with human judgment while maintaining the integrity of the evaluation process. By combining cosine similarity and brevity adjustment, we achieve a robust metric that accurately reflects both the content and quality of AI-generated text.

 \subsection{Combining Metrics to Form the ABHINAW Score: A comprehensive evaluation metric}

In previous sections, we addressed the limitations of traditional precision metrics by incorporating cosine similarity to handle creative text arrangements and brevity adjustment to penalize redundant text. This section combines these metrics into a single, robust evaluation metric named the ABHINAW Score, ensuring a comprehensive assessment of text fidelity in AI-generated images.

The ABHINAW Score is calculated by averaging the final scores of multiple candidate texts. This approach ensures stability and reliability across different iterations of AI-generated text.
\\

\textbf{Mathematical formulation}

Let \( S_i \) be the final score for the \( i \)-th candidate text, which incorporates both cosine similarity and brevity adjustment. The \textbf{ABHINAW Score} \( S_{\text{ABHINAW}} \) is given by:

\[
S_{\text{ABHINAW}} = \frac{1}{k} \sum_{i=1}^{k} S_i
\]

where \( k \) is the total number of candidate texts.

\textbf{Example:}\\

Consider the following detailed example with multiple candidates:

\begin{enumerate}
    \item \textbf{Reference Text}: "the"
    \item \textbf{Candidate Strings}: "the the", "the", "the", "the the", "the"
\end{enumerate}

 \begin{itemize}
    \item \textbf{Candidate 1}: "the the"
    \[
    \text{cosine\_similarity} = 1, \quad \text{BA} = e^{1 - \frac{7}{3}} = e^{-1.33} \approx 0.2644
    \]
    \[
    S_1 = 1 \times 0.2644 \approx 0.2644
    \]

    \item \textbf{Candidate 2}: "the"
    \[
    \text{cosine\_similarity} = 1, \quad \text{BA} = 1
    \]
    \[
    S_2 = 1 \times 1 = 1
    \]

    \item \textbf{Candidate 3}: "the"
    \[
    \text{cosine\_similarity} = 1, \quad \text{BA} = 1
    \]
    \[
    S_3 = 1 \times 1 = 1
    \]

    \item \textbf{Candidate 4}: "the the"
    \[
    \text{cosine\_similarity} = 1, \quad \text{BA} = e^{-1.33} \approx 0.2644
    \]
    \[
    S_4 = 1 \times 0.2644 \approx 0.2644
    \]

    \item \textbf{Candidate 5}: "the"
    \[
    \text{cosine\_similarity} = 1, \quad \text{BA} = 1
    \]
    \[
    S_5 = 1 \times 1 = 1
    \]
\end{itemize}

\textbf{Calculating the ABHINAW Score:}

\[
S_{\text{ABHINAW}} = \frac{1}{5} \sum_{i=1}^{5} S_i = \frac{1}{5} (0.2644 + 1 + 1 + 0.2644 + 1) = \frac{1}{5} \times 3.5288 \approx 0.7057
\]

\paragraph{\textbf{Overcoming initial precision problems}}

By integrating cosine similarity and brevity adjustment, we have systematically addressed the shortcomings of traditional precision metrics:

\begin{enumerate}
    \item \textbf{Inconsistency across generations}: Averaging the scores over multiple candidate texts ensures a stable and reliable evaluation metric.
    \item \textbf{Inadequacy in addressing error significance}: Using cosine similarity ensures that contextually similar texts are recognized, even if the arrangement of words is different.
    \item \textbf{Handling text arrangement}: Cosine similarity captures the similarity between texts irrespective of word order, addressing creative text arrangements effectively.
    \item \textbf{Penalizing redundancy}: Brevity adjustment penalizes over-generated text, ensuring that redundant texts do not receive high scores.
\end{enumerate}

By combining these approaches, the ABHINAW Score provides a comprehensive and robust metric for evaluating text fidelity in AI-generated images, aligning closely with human judgment and fostering greater creative freedom in AI image generation.

This comprehensive approach ensures that the ABHINAW Score accurately reflects both the content and quality of AI-generated text, making it a valuable tool for evaluating text Accuracy in AI-generated images.

\section{Experimental setup}

To systematically investigate the text accuracy in AI-generated images, multiple trials were conducted using DALL·E 3. For each trial, a series of images (10 images in our case) were generated based on consistent text prompts. The precision of the text within these images was evaluated using a predefined accuracy matrix, resulting in a precision score for each image. 

\begin{figure}[h!]
\centering
\includegraphics[width=0.9\textwidth]{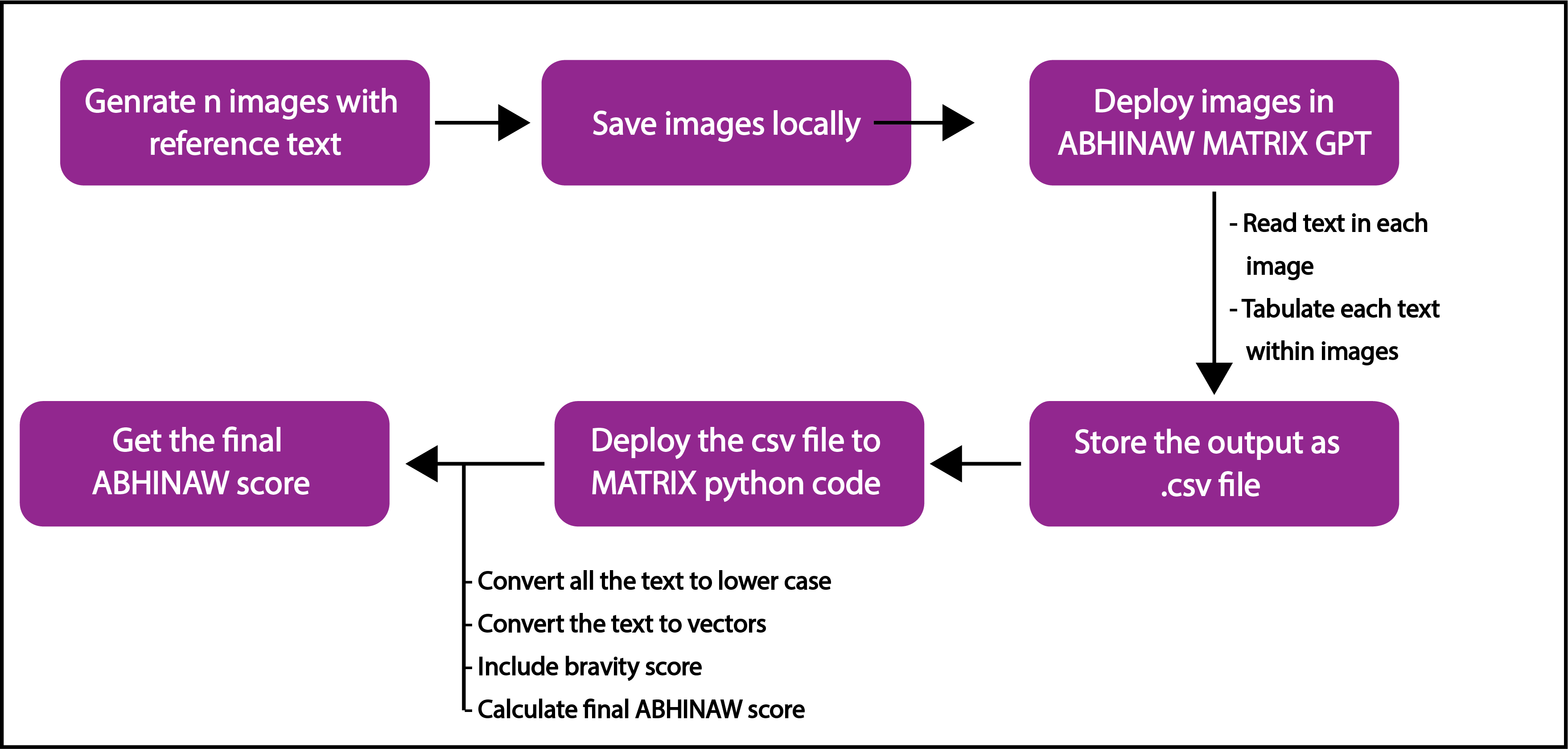}
\caption{Illustration of the methodology used in this paper}
\label{fig:steps}
\end{figure}

First, the AI generated images (ten images with each of the reference text) are generated using DALL E 3 provided with the reference text, these images were then stored locally. Next step is to evaluate weather the generated images have the reference text or not. In this case we have not used any traditional OCR, we have developed our own OCR using OpenAI Vision, this is by far accurate compared to other viable tools~\cite{shi2023exploring}. Our OCR is available as 'ABHINAW MATRIX GPT' in the GPT store of OpenAI. The ABHINAW MATRIX GPT is coded such that, any user just have to add the imaged to it and then the GPT will tabulate each text within the AI generated images. To eliminate the possibility of any misreading by our GPT, the previous step is repeated 10 times such that the the user will have now 100 tabulated text. These 100 text were stored locally in .csv file for further evaluation. In the next step, we have developed a mathematical framework where finally we are calculating the ABHINAW Score here are the main highlights of the framework, the stored text were converted to lower case (reason for incorporating the lower case is already discussed in section~\ref{subsec:Case-Sensitive precision inadequacy}). Then the generated text were converted to vectors and we calculate the cosine similarity for all the generated text (see section~\ref{subsec:Lack of Consideration for Text Arrangement}). If the cosine similarity is greater than 0.9 then only the text rearrangements are permitted in the scoring. In the final step, the text redundancy is taken into account by introduction of brevity adjustment (See section~\ref{subsec:brevity_adjustment}). The scores were then averaged to provide a normalized accuracy measure.
The detailed steps are illustrated in the figure~\ref{fig:steps}.

\section{Results}

\subsection{Accuracy of incorporated text within the AI generated image tends to decrease as the length of text increases}

ABHINAW score is used to evaluate the accuracy of text within the AI generated images. 
This experiment was done under two different scenarios, in one we have used English text containing 1 to 10 characters of frequently used words such as I, At, The, Line etc, in another scenario we have used English text containing 1-10 random characters. 
The exact words used for the study is mentioned in table~\ref{tab:ref_txt_known_unknown}. 

\begin{table}[h!]
\centering
\begin{tabular}{|c|c|c|}
\hline
\textbf{Length of text} & \textbf{Known Text} & \textbf{Unknown Text} \\
\hline
1 & i & v \\
2 & at & zl \\
3 & the & hua \\
4 & Line & sfbj \\
5 & Fruit & fzrsw \\
6 & Tables & miwwee \\
7 & hundred & ymbkgrc \\
8 & thousand & hfmhluxh \\
9 & Knowledge & abllcvisx \\
10 & basketball & csudcatayv \\
\hline
\end{tabular}
\caption{Reference text from Known and unknown categories used in this study}
\label{tab:ref_txt_known_unknown}
\end{table}

As mentioned in the experimental setup, each of the text (known and unknown) is subjected to reference text that needs to be incorporated within the AI generated images. 
Then each of these incorporated text is evaluated using ABHINAW Score. 
We observe that the score tends to decrease as the length of reference text is decreasing in case of known words as well as in case of unknown words. 
When observed closely, we noted that, in both the scenario once the threshold of six characters are crossed then there is rapid decrease in the accuracy score. 
The exact ABHINAW Score for the known and unknown words can be depicted from table~\ref{tab:abhinaw_scores}. 

\begin{table}[h!]
\centering
\begin{tabular}{|c|c|c|}
\hline
\multirow{2}{*}{\textbf{Text length}} & \multicolumn{2}{c|}{\textbf{ABHINAW Score (Automated)}} \\
\cline{2-3}
 & \textbf{Known Text} & \textbf{Unknown Text} \\
\hline
1 & 0.94 & 1 \\
2 & 0.99 & 0.86 \\
3 & 1 & 0.99 \\
4 & 0.98 & 0.76 \\
5 & 0.93 & 0.99 \\
6 & 0.92 & 0.97 \\
7 & 0.83 & 0.31 \\
8 & 0.65 & 0.48 \\
9 & 0.77 & 0.31 \\
10 & 0.79 & 0.34 \\
\hline
\end{tabular}
\caption{ABHINAW score for known and unknown texts by text length.}
\label{tab:abhinaw_scores}
\end{table}

\subsection{Manual evaluation of inserted text within the AI generated image are consistent with our automated evaluation}

In our previous experiments, we observe that, the ABHINAW score tends to decrease as the number of text within the reference text (known as well as unknown) increases. 
We wanted to ensure the efficacy of the vision tool of the OpenAI that we used to read the generated text. 
We decided to manually calculate the score by reading each of the generated text and calculate the ABHINAW matrix, for this particular experiment we have reduced the number of images that is to be evaluated to ten from each reference text. 
The exact ABHINAW Score for the known and Unknown reference text can be depicted from table~\ref{tab:comp_by_length}. 

\begin{figure}[h!]
\centering
\includegraphics[scale=0.8]{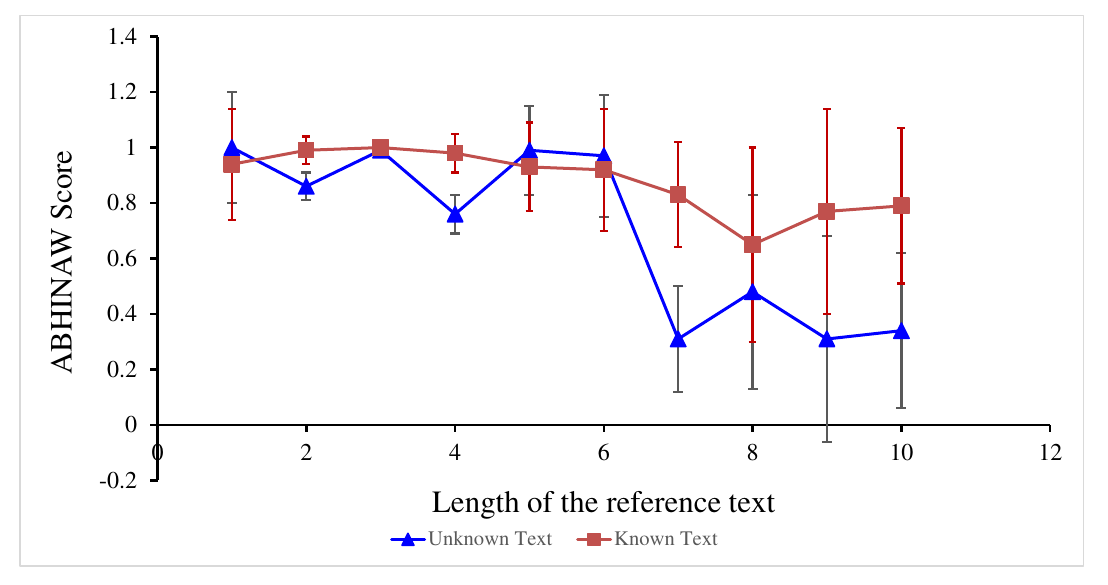}
\caption{Regression analysis on known texts as the length of text increases.}
\label{fig:reg_ana_known}
\end{figure}

We observed that, manual evaluation does not change our previous observation and we found that the efficiency of AI tolls decreases as the number of text in the reference text decreases (Figure~\ref{fig:reg_ana_known}). 
To ensure the similarity amongst the manual as well as automated evaluation of the ABHINAW score for known and unknown text, a statistical study must be performed. This study is presented in the next section. 

\begin{table}[h!]
\centering
\begin{tabular}{|c|c|c|}
\hline
\multirow{2}{*}{\textbf{Text length}} & \multicolumn{2}{c|}{\textbf{ABHINAW Score (Manual)}} \\
\cline{2-3}
 & \textbf{Known Text} & \textbf{Unknown Text} \\
\hline
1 & 0.86 & 1 \\
2 & 1 & 0.85 \\
3 & 1 & 1 \\
4 & 0.96 & 0.8 \\
5 & 0.91 & 0.91 \\
6 & 1 & 0.97 \\
7 & 0.83 & 0.32 \\
8 & 0.63 & 0.42 \\
9 & 0.75 & 0.18 \\
10 & 0.01 & 0.37 \\
\hline
\end{tabular}
\caption{Comparison of ABHINAW score calculated manually for known and unknown texts with respect to text length.}
\label{tab:comp_by_length}
\end{table}

\subsection{Correlation study for the automated and manual evaluation for incorporated text within the AI generated imaged shows a positive correlation}

\begin{figure}[h!]
\centering
\includegraphics[width=0.8\textwidth]{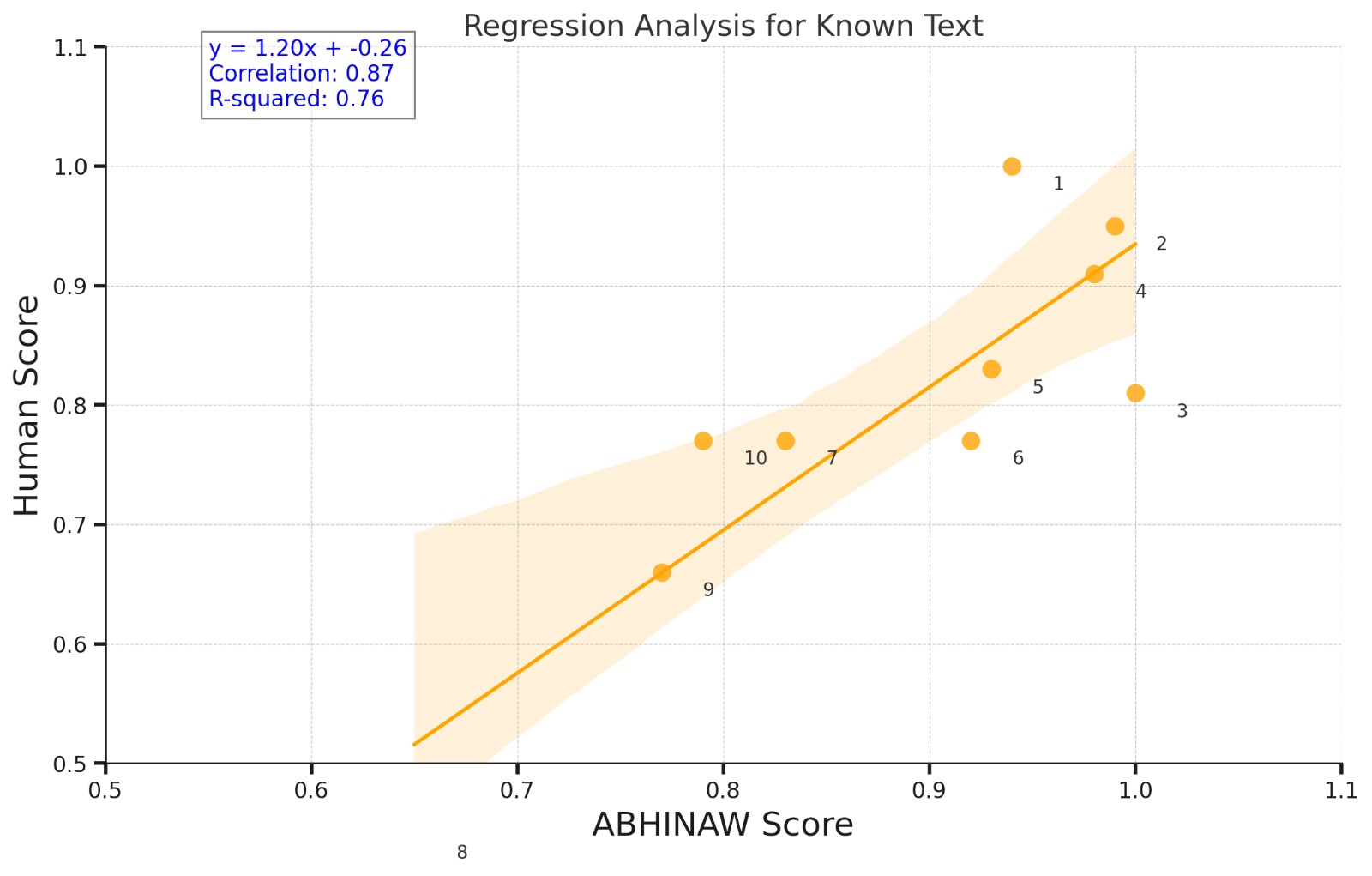}
\caption{Regression line comparing human and automated evaluations for known texts as the length of text increases.}
\label{fig:reg_known}
\end{figure}

\begin{figure}[h!]
\centering
\includegraphics[width=0.8\textwidth]{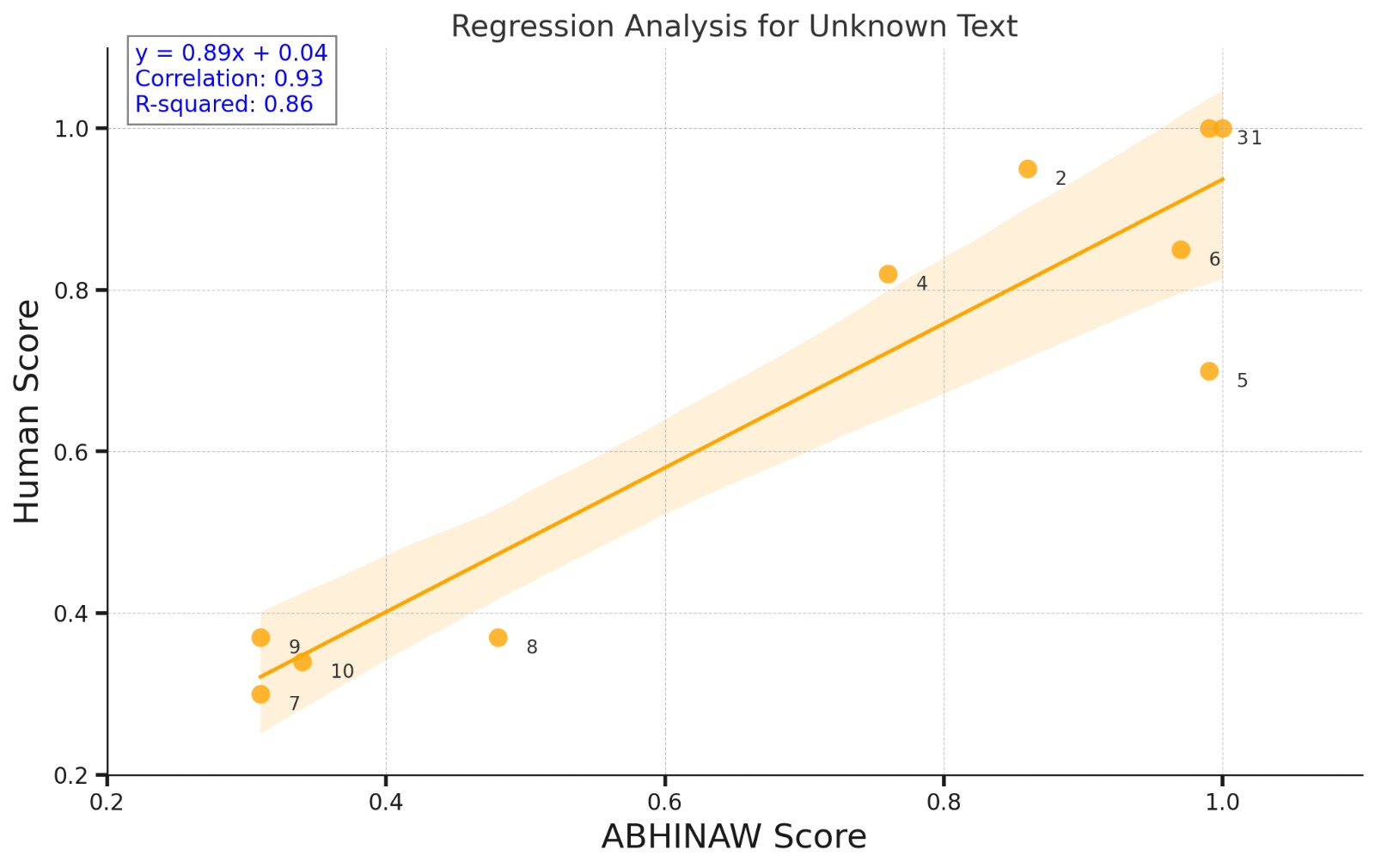}
\caption{Regression line comparing human and automated evaluations for unknown texts as the length of text increases.}
\label{fig:reg_unknown}
\end{figure}

In this analysis we have performed the regression analysis between the automated and manually evaluated ABHINAW scores for the incorporated text within the AI generated images. This study is performed for known as well as unknown texts. We observed a positive correlation exist with permissible goodness of fit for both cases (Figure~\ref{fig:reg_known} for known text and Figure~\ref{fig:reg_unknown} for unknown text). This ensures that our automated method to find out the efficacy of incorporated text within the AI generated images is as good as the manually calculated method.  

\section{Conclusion and future scope}
In this study we have introduced the ABHINAW Score, a comprehensive metric for evaluating text fidelity in AI-generated images. After discussion various aspects of text incoporation of within the AI generated imaged we fairly discuss the need of evaluation matrix for the comparison of output and what user actually wanted. We also discussed various corners of this matrix which needs to be taken care of such as creative freedom of the AI, case sensitivity, text arrangements etc. We have introduced several methods to counter these problems by integrating cosine similarity and brevity adjustment and come up with ABHINAW Score that represent the efficacy of the AI system to generate imaged that are much closer to the hypothetical image with exact reference text. This metric addresses both creative text arrangements and redundancy, providing a robust evaluation framework. 
The experimental results, supported by statistical analysis, demonstrate a strong correlation between human and automated evaluations, validating the effectiveness of the ABHINAW Score.

Anticipating future advancements in text generation within AI-generated images, we propose several enhancements and potential research directions:

\textit{Two-Prompt Solution:} A promising future direction is the implementation of the "two-prompt solution." 
This innovative approach suggests incorporating two distinct text input fields in image generation tools: \textit{(i) General Description Prompt:} The first input field will continue to describe the scene or provide general image descriptions, similar to current single-prompt systems.
\textit{(ii) Text Specification Prompt:} The second input field will be dedicated exclusively to specifying the exact text to be rendered in the image. 
This ensures that only the text entered in this field, known as "prompt 2," appears in the generated image.
The dual-prompt system aims to eliminate ambiguity about the text's intended appearance, enhancing the precision of text rendering in AI-generated images. 
While full implementation may take time, this approach promises significant improvements in text accuracy.

\begin{figure}[h]
    \centering
    \includegraphics[scale=0.3]{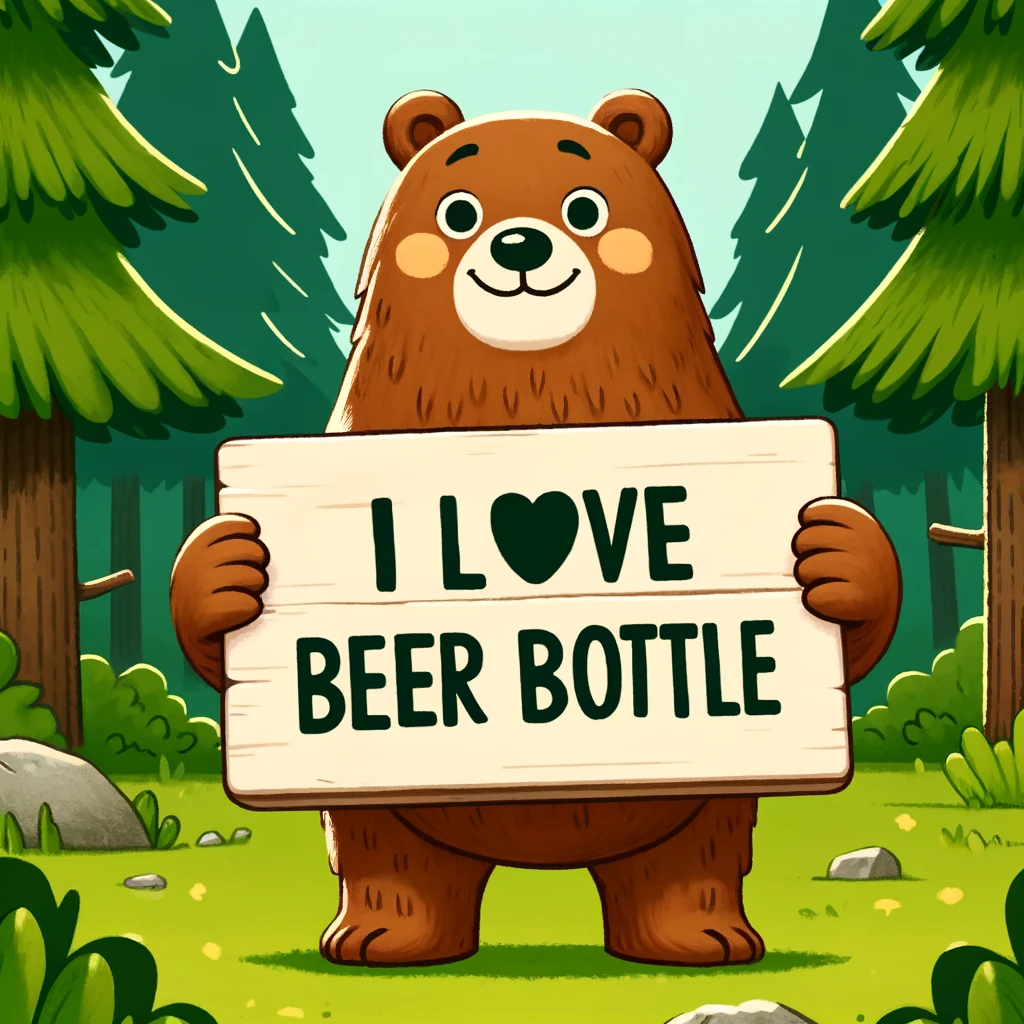}
    \caption{Need of visual cosine similarity}
    \label{fig:vc}
\end{figure}

\textit{Dynamic Background Regeneration:} We hypothesize that Dynamic Background Regeneration could serve as a novel method to enhance text generation in AI-generated images. Building on existing technologies like ControlNet, which effectively controls image structure, this approach would involve a dual-layer generation process. One layer would generate the background, while a second layer would focus on text placement.

In this proposed method, regions designated for text would undergo zero convolutions in the background layer, potentially enabling seamless integration. A key feature of this hypothesis is that the text layer would remain editable post-generation, allowing for corrections and transformations. As text modifications are made, the background would dynamically regenerate to maintain visual coherence.

If developed, this approach could offer a flexible solution for precise and adaptable text generation, potentially improving the usability of AI-generated images across various applications.

\textit{Addressing the Scope of the Evaluation Matrix:} One potential criticism is that the evaluation matrix might appear overly focused on text accuracy, potentially leading to scenarios where simply outputting text in tools like Microsoft Word or Paint could yield seemingly perfect results. 
However, it is crucial to clarify that the accuracy of the ABHINAW Score is specifically designed for AI-generated images.\\ 

\textit{Visual Cosine Similarity :} Future research should address the problem of visual cosine similarity. 
For example, the system should be able to recognize that the output ``4" and ``four" or a heart symbol instead of the word ``love" should be considered similar.As shown in Figure~\ref{fig:vc}.
This enhancement will improve the evaluation of text fidelity by recognizing visually similar representations.

The proposed enhancements will contribute to the ongoing development of more accurate and reliable methods for evaluating text in AI-generated images. 
By addressing these future challenges, the ABHINAW Score and its associated methodologies can be refined to provide even greater precision and reliability in text evaluation.

\bibliography{reference}

\begin{thebibliography}{68}
\providecommand{\natexlab}[1]{#1}
\providecommand{\url}[1]{\texttt{#1}}
\expandafter\ifx\csname urlstyle\endcsname\relax
  \providecommand{\doi}[1]{doi: #1}\else
  \providecommand{\doi}{doi: \begingroup \urlstyle{rm}\Url}\fi

\bibitem[mid(2024)]{midjourney}
Midjourney, 2024.
\newblock URL \url{https://www.midjourney.com/home}.
\newblock Accessed: 2024-07-15.

\bibitem[Bakr et~al.(2023)Bakr, Sun, Shen, Khan, Li, and Elhoseiny]{bakr2023hrs}
Eslam~Mohamed Bakr, Pengzhan Sun, Xiaoqian Shen, Faizan~Farooq Khan, Li~Erran Li, and Mohamed Elhoseiny.
\newblock Hrs-bench: Holistic, reliable and scalable benchmark for text-to-image models.
\newblock \emph{arXiv preprint arXiv:2304.05390}, 2023.

\bibitem[Betker et~al.(2023)Betker, Goh, Jing, Brooks, Wang, Li, Ouyang, Zhuang, Lee, Guo, et~al.]{betker2023improving}
James Betker, Gabriel Goh, Li~Jing, Tim Brooks, Jianfeng Wang, Linjie Li, Long Ouyang, Juntang Zhuang, Joyce Lee, Yufei Guo, et~al.
\newblock Improving image generation with better captions.
\newblock \emph{Computer Science. https://cdn. openai. com/papers/dall-e-3. pdf}, 2\penalty0 (3):\penalty0 8, 2023.

\bibitem[Chefer et~al.(2023)Chefer, Alaluf, Vinker, Wolf, and Cohen-Or]{chefer2023attend}
Hila Chefer, Yuval Alaluf, Yael Vinker, Lior Wolf, and Daniel Cohen-Or.
\newblock Attend-and-excite: Attention-based semantic guidance for text-to-image diffusion models.
\newblock 2023.

\bibitem[Chen et~al.(2023{\natexlab{a}})Chen, Yu, Ge, Yao, Xie, Wu, Wang, Kwok, Luo, Lu, et~al.]{chen2023pixart}
Junsong Chen, Jincheng Yu, Chongjian Ge, Lewei Yao, Enze Xie, Yue Wu, Zhongdao Wang, James Kwok, Ping Luo, Huchuan Lu, et~al.
\newblock Pixart-$\alpha$: Fast training of diffusion transformer for photorealistic text-to-image synthesis.
\newblock \emph{arXiv preprint arXiv:2310.00426}, 2023{\natexlab{a}}.

\bibitem[Chen et~al.(2023{\natexlab{b}})Chen, Laina, and Vedaldi]{chen2023trainingfree}
Minghao Chen, Iro Laina, and Andrea Vedaldi.
\newblock Training-free layout control with cross-attention guidance.
\newblock \emph{arXiv preprint arXiv:2304.03373}, 2023{\natexlab{b}}.

\bibitem[Chen et~al.(2023{\natexlab{c}})Chen, Liu, Yang, Yuan, You, Liu, and Yang]{chen2023reason}
Xiaohui Chen, Yongfei Liu, Yingxiang Yang, Jianbo Yuan, Quanzeng You, Li-Ping Liu, and Hongxia Yang.
\newblock Reason out your layout: Evoking the layout master from large language models for text-to-image synthesis.
\newblock \emph{arXiv preprint arXiv:2311.17126}, 2023{\natexlab{c}}.

\bibitem[Chen(2023)]{chen2023xiqe}
Yixiong Chen.
\newblock X-iqe: explainable image quality evaluation for text-to-image generation with visual large language models.
\newblock \emph{arXiv preprint arXiv:2305.10843}, 2023.

\bibitem[Cho et~al.(2022)Cho, Zala, and Bansal]{cho2022dall}
Jaemin Cho, Abhay Zala, and Mohit Bansal.
\newblock Dall-eval: Probing the reasoning skills and social biases of text-to-image generative transformers.
\newblock \emph{arXiv preprint arXiv:2202.04053}, 2022.

\bibitem[Dong et~al.(2023)Dong, Xiong, Goyal, Pan, Diao, Zhang, Shum, and Zhang]{dong2023raft}
Hanze Dong, Wei Xiong, Deepanshu Goyal, Rui Pan, Shizhe Diao, Jipeng Zhang, Kashun Shum, and Tong Zhang.
\newblock Raft: Reward ranked finetuning for generative foundation model alignment.
\newblock \emph{arXiv preprint arXiv:2304.06767}, 2023.

\bibitem[Esser et~al.(2024)Esser, Kulal, Blattmann, Entezari, M{\"u}ller, Saini, Levi, Lorenz, Sauer, Boesel, et~al.]{esser2024scaling}
Patrick Esser, Sumith Kulal, Andreas Blattmann, Rahim Entezari, Jonas M{\"u}ller, Harry Saini, Yam Levi, Dominik Lorenz, Axel Sauer, Frederic Boesel, et~al.
\newblock Scaling rectified flow transformers for high-resolution image synthesis.
\newblock In \emph{Forty-first International Conference on Machine Learning}, 2024.

\bibitem[Feng et~al.(2023)Feng, He, Fu, Jampani, Akula, Narayana, Basu, Wang, and Wang]{feng2022training}
Weixi Feng, Xuehai He, Tsu-Jui Fu, Varun Jampani, Arjun Akula, Pradyumna Narayana, Sugato Basu, Xin~Eric Wang, and William~Yang Wang.
\newblock Training-free structured diffusion guidance for compositional text-to-image synthesis.
\newblock In \emph{ICLR}, 2023.

\bibitem[Gafni et~al.(2022)Gafni, Polyak, Ashual, Sheynin, Parikh, and Taigman]{gafni2022make}
Oran Gafni, Adam Polyak, Oron Ashual, Shelly Sheynin, Devi Parikh, and Yaniv Taigman.
\newblock Make-a-scene: Scene-based text-to-image generation with human priors.
\newblock In \emph{ECCV}, 2022.

\bibitem[Gani et~al.(2023)Gani, Bhat, Naseer, Khan, and Wonka]{gani2023llm}
Hanan Gani, Shariq~Farooq Bhat, Muzammal Naseer, Salman Khan, and Peter Wonka.
\newblock Llm blueprint: Enabling text-to-image generation with complex and detailed prompts.
\newblock \emph{arXiv preprint arXiv:2310.10640}, 2023.

\bibitem[Goodfellow et~al.(2014)Goodfellow, Pouget-Abadie, Mirza, Xu, Warde-Farley, Ozair, Courville, and Bengio]{goodfellow2014generative}
Ian Goodfellow, Jean Pouget-Abadie, Mehdi Mirza, Bing Xu, David Warde-Farley, Sherjil Ozair, Aaron Courville, and Yoshua Bengio.
\newblock Generative adversarial nets.
\newblock 2014.

\bibitem[Gu et~al.(2023)Gu, Zhai, Zhang, Susskind, and Jaitly]{gu2023matryoshka}
Jiatao Gu, Shuangfei Zhai, Yizhe Zhang, Joshua~M Susskind, and Navdeep Jaitly.
\newblock Matryoshka diffusion models.
\newblock In \emph{The Twelfth International Conference on Learning Representations}, 2023.

\bibitem[Heusel et~al.(2017)Heusel, Ramsauer, Unterthiner, Nessler, and Hochreiter]{heusel2018gans}
Martin Heusel, Hubert Ramsauer, Thomas Unterthiner, Bernhard Nessler, and Sepp Hochreiter.
\newblock Gans trained by a two time-scale update rule converge to a local nash equilibrium, 2017.

\bibitem[Huynh-Thu and Ghanbari(2008)]{huynh2008scope}
Quan Huynh-Thu and Mohammed Ghanbari.
\newblock Scope of validity of psnr in image/video quality assessment.
\newblock \emph{Electronics letters}, 44\penalty0 (13):\penalty0 800--801, 2008.

\bibitem[Kim et~al.(2023)Kim, Lee, Kim, Ha, and Zhu]{kim2023dense}
Yunji Kim, Jiyoung Lee, Jin-Hwa Kim, Jung-Woo Ha, and Jun-Yan Zhu.
\newblock Dense text-to-image generation with attention modulation.
\newblock In \emph{Proceedings of the IEEE/CVF International Conference on Computer Vision}, pages 7701--7711, 2023.

\bibitem[Kirstain et~al.(2023)Kirstain, Polyak, Singer, Matiana, Penna, and Levy]{kirstain2023pick}
Yuval Kirstain, Adam Polyak, Uriel Singer, Shahbuland Matiana, Joe Penna, and Omer Levy.
\newblock Pick-a-pic: An open dataset of user preferences for text-to-image generation.
\newblock \emph{Advances in Neural Information Processing Systems}, 36:\penalty0 36652--36663, 2023.

\bibitem[Ku et~al.(2023{\natexlab{a}})Ku, Jiang, Wei, Yue, and Chen]{ku2023viescore}
Max Ku, Dongfu Jiang, Cong Wei, Xiang Yue, and Wenhu Chen.
\newblock Viescore: Towards explainable metrics for conditional image synthesis evaluation.
\newblock \emph{arXiv preprint arXiv:2312.14867}, 2023{\natexlab{a}}.

\bibitem[Ku et~al.(2023{\natexlab{b}})Ku, Li, Zhang, Lu, Fu, Zhuang, and Chen]{ku2023imagenhub}
Max Ku, Tianle Li, Kai Zhang, Yujie Lu, Xingyu Fu, Wenwen Zhuang, and Wenhu Chen.
\newblock Imagenhub: Standardizing the evaluation of conditional image generation models.
\newblock \emph{arXiv preprint arXiv:2310.01596}, 2023{\natexlab{b}}.

\bibitem[Lee et~al.(2023)Lee, Liu, Ryu, Watkins, Du, Boutilier, Abbeel, Ghavamzadeh, and Gu]{lee2023aligning}
Kimin Lee, Hao Liu, Moonkyung Ryu, Olivia Watkins, Yuqing Du, Craig Boutilier, Pieter Abbeel, Mohammad Ghavamzadeh, and Shixiang~Shane Gu.
\newblock Aligning text-to-image models using human feedback.
\newblock \emph{arXiv preprint arXiv:2302.12192}, 2023.

\bibitem[Lee et~al.(2024)Lee, Yasunaga, Meng, Mai, Park, Gupta, Zhang, Narayanan, Teufel, Bellagente, et~al.]{lee2024holistic}
Tony Lee, Michihiro Yasunaga, Chenlin Meng, Yifan Mai, Joon~Sung Park, Agrim Gupta, Yunzhi Zhang, Deepak Narayanan, Hannah Teufel, Marco Bellagente, et~al.
\newblock Holistic evaluation of text-to-image models.
\newblock \emph{Advances in Neural Information Processing Systems}, 36, 2024.

\bibitem[Li et~al.(2022{\natexlab{a}})Li, Li, Xiong, and Hoi]{li2022blip}
Junnan Li, Dongxu Li, Caiming Xiong, and Steven Hoi.
\newblock Blip: Bootstrapping language-image pre-training for unified vision-language understanding and generation.
\newblock In \emph{International Conference on Machine Learning}, pages 12888--12900. PMLR, 2022{\natexlab{a}}.

\bibitem[Li et~al.(2023{\natexlab{a}})Li, Li, Savarese, and Hoi]{li2023blip}
Junnan Li, Dongxu Li, Silvio Savarese, and Steven Hoi.
\newblock Blip-2: Bootstrapping language-image pre-training with frozen image encoders and large language models.
\newblock \emph{arXiv preprint arXiv:2301.12597}, 2023{\natexlab{a}}.

\bibitem[Li et~al.(2023{\natexlab{b}})Li, Liu, Wu, Mu, Yang, Gao, Li, and Lee]{li2023gligen}
Yuheng Li, Haotian Liu, Qingyang Wu, Fangzhou Mu, Jianwei Yang, Jianfeng Gao, Chunyuan Li, and Yong~Jae Lee.
\newblock Gligen: Open-set grounded text-to-image generation.
\newblock In \emph{Proceedings of the IEEE/CVF Conference on Computer Vision and Pattern Recognition}, pages 22511--22521, 2023{\natexlab{b}}.

\bibitem[Li et~al.(2022{\natexlab{b}})Li, Min, Li, and Xu]{li2022stylet2i}
Zhiheng Li, Martin~Renqiang Min, Kai Li, and Chenliang Xu.
\newblock Stylet2i: Toward compositional and high-fidelity text-to-image synthesis.
\newblock In \emph{Proceedings of the IEEE/CVF Conference on Computer Vision and Pattern Recognition}, pages 18197--18207, 2022{\natexlab{b}}.

\bibitem[Lian et~al.(2023)Lian, Li, Yala, and Darrell]{lian2023llm}
Long Lian, Boyi Li, Adam Yala, and Trevor Darrell.
\newblock Llm-grounded diffusion: Enhancing prompt understanding of text-to-image diffusion models with large language models.
\newblock \emph{arXiv preprint arXiv:2305.13655}, 2023.

\bibitem[Liang et~al.(2024)Liang, He, Li, Li, Klimovskiy, Carolan, Sun, Pont-Tuset, Young, Yang, et~al.]{liang2024rich}
Youwei Liang, Junfeng He, Gang Li, Peizhao Li, Arseniy Klimovskiy, Nicholas Carolan, Jiao Sun, Jordi Pont-Tuset, Sarah Young, Feng Yang, et~al.
\newblock Rich human feedback for text-to-image generation.
\newblock In \emph{Proceedings of the IEEE/CVF Conference on Computer Vision and Pattern Recognition}, pages 19401--19411, 2024.

\bibitem[Lin et~al.(2014)Lin, Maire, Belongie, Hays, Perona, Ramanan, Doll{\'a}r, and Zitnick]{lin2014microsoft}
Tsung-Yi Lin, Michael Maire, Serge Belongie, James Hays, Pietro Perona, Deva Ramanan, Piotr Doll{\'a}r, and C~Lawrence Zitnick.
\newblock Microsoft coco: Common objects in context.
\newblock In \emph{Computer Vision--ECCV 2014: 13th European Conference, Zurich, Switzerland, September 6-12, 2014, Proceedings, Part V 13}, pages 740--755. Springer, 2014.

\bibitem[Liu et~al.(2022)Liu, Li, Du, Torralba, and Tenenbaum]{liu2022compositional}
Nan Liu, Shuang Li, Yilun Du, Antonio Torralba, and Joshua~B Tenenbaum.
\newblock Compositional visual generation with composable diffusion models.
\newblock In \emph{Computer Vision--ECCV 2022: 17th European Conference, Tel Aviv, Israel, October 23--27, 2022, Proceedings, Part XVII}, pages 423--439. Springer, 2022.

\bibitem[Lu et~al.(2023)Lu, Yang, Li, Wang, and Wang]{lu2023llmscore}
Yujie Lu, Xianjun Yang, Xiujun Li, Xin~Eric Wang, and William~Yang Wang.
\newblock Llmscore: Unveiling the power of large language models in text-to-image synthesis evaluation.
\newblock \emph{arXiv preprint arXiv:2305.11116}, 2023.

\bibitem[Meral et~al.(2024)Meral, Simsar, Tombari, and Yanardag]{meral2024conform}
Tuna Han~Salih Meral, Enis Simsar, Federico Tombari, and Pinar Yanardag.
\newblock Conform: Contrast is all you need for high-fidelity text-to-image diffusion models.
\newblock In \emph{Proceedings of the IEEE/CVF Conference on Computer Vision and Pattern Recognition}, pages 9005--9014, 2024.

\bibitem[Nichol et~al.(2022)Nichol, Dhariwal, Ramesh, Shyam, Mishkin, McGrew, Sutskever, and Chen]{nichol2022glide}
Alex Nichol, Prafulla Dhariwal, Aditya Ramesh, Pranav Shyam, Pamela Mishkin, Bob McGrew, Ilya Sutskever, and Mark Chen.
\newblock Glide: Towards photorealistic image generation and editing with text-guided diffusion models.
\newblock In \emph{ICML}, 2022.

\bibitem[Nilsback and Zisserman(2008)]{nilsback2008automated}
Maria-Elena Nilsback and Andrew Zisserman.
\newblock Automated flower classification over a large number of classes.
\newblock In \emph{2008 Sixth Indian Conference on Computer Vision, Graphics \& Image Processing}, pages 722--729. IEEE, 2008.

\bibitem[Papineni et~al.(2002)Papineni, Roukos, Ward, and Zhu]{papineni-etal-2002-bleu}
Kishore Papineni, Salim Roukos, Todd Ward, and Wei-Jing Zhu.
\newblock {B}leu: a method for automatic evaluation of machine translation.
\newblock In Pierre Isabelle, Eugene Charniak, and Dekang Lin, editors, \emph{Proceedings of the 40th Annual Meeting of the Association for Computational Linguistics}, pages 311--318, Philadelphia, Pennsylvania, USA, July 2002. Association for Computational Linguistics.
\newblock \doi{10.3115/1073083.1073135}.
\newblock URL \url{https://aclanthology.org/P02-1040}.

\bibitem[Park et~al.(2021)Park, Azadi, Liu, Darrell, and Rohrbach]{park2021benchmark}
Dong~Huk Park, Samaneh Azadi, Xihui Liu, Trevor Darrell, and Anna Rohrbach.
\newblock Benchmark for compositional text-to-image synthesis.
\newblock In \emph{Thirty-fifth Conference on Neural Information Processing Systems Datasets and Benchmarks Track (Round 1)}, 2021.

\bibitem[Petsiuk et~al.(2022)Petsiuk, Siemenn, Surbehera, Chin, Tyser, Hunter, Raghavan, Hicke, Plummer, Kerret, et~al.]{petsiuk2022human}
Vitali Petsiuk, Alexander~E Siemenn, Saisamrit Surbehera, Zad Chin, Keith Tyser, Gregory Hunter, Arvind Raghavan, Yann Hicke, Bryan~A Plummer, Ori Kerret, et~al.
\newblock Human evaluation of text-to-image models on a multi-task benchmark.
\newblock \emph{arXiv preprint arXiv:2211.12112}, 2022.

\bibitem[Podell et~al.(2023)Podell, English, Lacey, Blattmann, Dockhorn, M{\"u}ller, Penna, and Rombach]{podell2023sdxl}
Dustin Podell, Zion English, Kyle Lacey, Andreas Blattmann, Tim Dockhorn, Jonas M{\"u}ller, Joe Penna, and Robin Rombach.
\newblock Sdxl: Improving latent diffusion models for high-resolution image synthesis.
\newblock \emph{arXiv preprint arXiv:2307.01952}, 2023.

\bibitem[Radford et~al.(2021)Radford, Kim, Hallacy, Ramesh, Goh, Agarwal, Sastry, Askell, Mishkin, Clark, et~al.]{radford2021learning}
Alec Radford, Jong~Wook Kim, Chris Hallacy, Aditya Ramesh, Gabriel Goh, Sandhini Agarwal, Girish Sastry, Amanda Askell, Pamela Mishkin, Jack Clark, et~al.
\newblock Learning transferable visual models from natural language supervision.
\newblock In \emph{International conference on machine learning}, pages 8748--8763. PMLR, 2021.

\bibitem[Ramesh et~al.(2021)Ramesh, Pavlov, Goh, Gray, Voss, Radford, Chen, and Sutskever]{ramesh2021zero}
Aditya Ramesh, Mikhail Pavlov, Gabriel Goh, Scott Gray, Chelsea Voss, Alec Radford, Mark Chen, and Ilya Sutskever.
\newblock Zero-shot text-to-image generation.
\newblock In \emph{ICML}, 2021.

\bibitem[Ramesh et~al.(2022{\natexlab{a}})Ramesh, Dhariwal, Nichol, Chu, and Chen]{ramesh2022hierarchical}
Aditya Ramesh, Prafulla Dhariwal, Alex Nichol, Casey Chu, and Mark Chen.
\newblock Hierarchical text-conditional image generation with clip latents.
\newblock \emph{arXiv preprint arXiv:2204.06125}, 2022{\natexlab{a}}.

\bibitem[Ramesh et~al.(2022{\natexlab{b}})Ramesh, Pavlov, Goh, Gray, Agarwal, Radford, Kim, and Mahajan]{ramesh2022dalle2}
Aditya Ramesh, Mikhail Pavlov, Gabriel Goh, Scott Gray, Chelsea~Voss Agarwal, Alec Radford, Jong~Wook Kim, and Prafulla Mahajan.
\newblock Dall-e 2: A new text-to-image generation model, 2022{\natexlab{b}}.
\newblock URL \url{https://cdn.openai.com/papers/dall-e-2.pdf}.
\newblock Accessed: 2024-07-15.

\bibitem[Rassin et~al.(2024)Rassin, Hirsch, Glickman, Ravfogel, Goldberg, and Chechik]{rassin2024linguistic}
Royi Rassin, Eran Hirsch, Daniel Glickman, Shauli Ravfogel, Yoav Goldberg, and Gal Chechik.
\newblock Linguistic binding in diffusion models: Enhancing attribute correspondence through attention map alignment.
\newblock \emph{Advances in Neural Information Processing Systems}, 36, 2024.

\bibitem[Reed et~al.(2016{\natexlab{a}})Reed, Akata, Yan, Logeswaran, Schiele, and Lee]{reed2016generative}
Scott Reed, Zeynep Akata, Xinchen Yan, Lajanugen Logeswaran, Bernt Schiele, and Honglak Lee.
\newblock Generative adversarial text to image synthesis.
\newblock In \emph{ICML}, 2016{\natexlab{a}}.

\bibitem[Reed et~al.(2016{\natexlab{b}})Reed, Akata, Mohan, Tenka, Schiele, and Lee]{reed2016learning}
Scott~E Reed, Zeynep Akata, Santosh Mohan, Samuel Tenka, Bernt Schiele, and Honglak Lee.
\newblock Learning what and where to draw.
\newblock In \emph{NeurIPS}, 2016{\natexlab{b}}.

\bibitem[Rombach et~al.(2022)Rombach, Blattmann, Lorenz, Esser, and Ommer]{rombach2022high}
Robin Rombach, Andreas Blattmann, Dominik Lorenz, Patrick Esser, and Bj{\"o}rn Ommer.
\newblock High-resolution image synthesis with latent diffusion models.
\newblock In \emph{Proceedings of the IEEE/CVF Conference on Computer Vision and Pattern Recognition}, pages 10684--10695, 2022.

\bibitem[Saharia et~al.(2022)Saharia, Chan, Saxena, Li, Whang, Denton, Ghasemipour, Gontijo~Lopes, Karagol~Ayan, Salimans, et~al.]{saharia2022imagen}
Chitwan Saharia, William Chan, Saurabh Saxena, Lala Li, Jay Whang, Emily~L Denton, Kamyar Ghasemipour, Raphael Gontijo~Lopes, Burcu Karagol~Ayan, Tim Salimans, et~al.
\newblock Photorealistic text-to-image diffusion models with deep language understanding.
\newblock \emph{Advances in Neural Information Processing Systems}, 35:\penalty0 36479--36494, 2022.

\bibitem[Salimans et~al.(2016)Salimans, Goodfellow, Zaremba, Cheung, Radford, and Chen]{salimans2016improved}
Tim Salimans, Ian Goodfellow, Wojciech Zaremba, Vicki Cheung, Alec Radford, and Xi~Chen.
\newblock Improved techniques for training gans.
\newblock In \emph{Proceedings of the 30th International Conference on Neural Information Processing Systems}, NIPS'16, page 2234–2242, Red Hook, NY, USA, 2016. Curran Associates Inc.
\newblock ISBN 9781510838819.

\bibitem[Shi et~al.(2023)Shi, Peng, Liao, Lin, Chen, Liu, Zhang, and Jin]{shi2023exploring}
Yongxin Shi, Dezhi Peng, Wenhui Liao, Zening Lin, Xinhong Chen, Chongyu Liu, Yuyi Zhang, and Lianwen Jin.
\newblock Exploring ocr capabilities of gpt-4v (ision): A quantitative and in-depth evaluation.
\newblock \emph{arXiv preprint arXiv:2310.16809}, 2023.

\bibitem[Sun et~al.(2023)Sun, Fu, Hu, Wang, Rassin, Juan, Alon, Herrmann, van Steenkiste, Krishna, et~al.]{sun2023dreamsync}
Jiao Sun, Deqing Fu, Yushi Hu, Su~Wang, Royi Rassin, Da-Cheng Juan, Dana Alon, Charles Herrmann, Sjoerd van Steenkiste, Ranjay Krishna, et~al.
\newblock Dreamsync: Aligning text-to-image generation with image understanding feedback.
\newblock In \emph{Synthetic Data for Computer Vision Workshop@ CVPR 2024}, 2023.

\bibitem[Taghipour et~al.(2024)Taghipour, Ghahremani, Bennamoun, Rekavandi, Laga, and Boussaid]{taghipour2024box}
Ashkan Taghipour, Morteza Ghahremani, Mohammed Bennamoun, Aref~Miri Rekavandi, Hamid Laga, and Farid Boussaid.
\newblock Box it to bind it: Unified layout control and attribute binding in t2i diffusion models.
\newblock \emph{arXiv preprint arXiv:2402.17910}, 2024.

\bibitem[Wah et~al.(2011)Wah, Branson, Welinder, Perona, and Belongie]{wah2011caltech}
Catherine Wah, Steve Branson, Peter Welinder, Pietro Perona, and Serge Belongie.
\newblock The caltech-ucsd birds-200-2011 dataset.
\newblock 2011.

\bibitem[Wang et~al.(2024{\natexlab{a}})Wang, Chen, Chen, Ma, Lu, and Lin]{wang2024compositional}
Ruichen Wang, Zekang Chen, Chen Chen, Jian Ma, Haonan Lu, and Xiaodong Lin.
\newblock Compositional text-to-image synthesis with attention map control of diffusion models.
\newblock In \emph{Proceedings of the AAAI Conference on Artificial Intelligence}, volume~38, pages 5544--5552, 2024{\natexlab{a}}.

\bibitem[Wang et~al.(2024{\natexlab{b}})Wang, Xie, Li, Wang, Liu, and Li]{wang2024divide}
Zhenyu Wang, Enze Xie, Aoxue Li, Zhongdao Wang, Xihui Liu, and Zhenguo Li.
\newblock Divide and conquer: Language models can plan and self-correct for compositional text-to-image generation.
\newblock \emph{arXiv preprint arXiv:2401.15688}, 2024{\natexlab{b}}.

\bibitem[Wu et~al.(2023{\natexlab{a}})Wu, Liu, Zhao, Bui, Lin, Zhang, and Chang]{wu2023harnessing}
Qiucheng Wu, Yujian Liu, Handong Zhao, Trung Bui, Zhe Lin, Yang Zhang, and Shiyu Chang.
\newblock Harnessing the spatial-temporal attention of diffusion models for high-fidelity text-to-image synthesis.
\newblock \emph{arXiv preprint arXiv:2304.03869}, 2023{\natexlab{a}}.

\bibitem[Wu et~al.(2023{\natexlab{b}})Wu, Hao, Sun, Chen, Zhu, Zhao, and Li]{wu2023human}
Xiaoshi Wu, Yiming Hao, Keqiang Sun, Yixiong Chen, Feng Zhu, Rui Zhao, and Hongsheng Li.
\newblock Human preference score v2: A solid benchmark for evaluating human preferences of text-to-image synthesis.
\newblock \emph{arXiv preprint arXiv:2306.09341}, 2023{\natexlab{b}}.

\bibitem[Wu et~al.(2023{\natexlab{c}})Wu, Sun, Zhu, Zhao, and Li]{wu2023better}
Xiaoshi Wu, Keqiang Sun, Feng Zhu, Rui Zhao, and Hongsheng Li.
\newblock Better aligning text-to-image models with human preference, 2023{\natexlab{c}}.

\bibitem[Xu et~al.(2024)Xu, Liu, Wu, Tong, Li, Ding, Tang, and Dong]{xu2024imagereward}
Jiazheng Xu, Xiao Liu, Yuchen Wu, Yuxuan Tong, Qinkai Li, Ming Ding, Jie Tang, and Yuxiao Dong.
\newblock Imagereward: Learning and evaluating human preferences for text-to-image generation.
\newblock \emph{Advances in Neural Information Processing Systems}, 36, 2024.

\bibitem[Xu et~al.(2018)Xu, Zhang, Huang, Zhang, Gan, Huang, and He]{xu2018attngan}
Tao Xu, Pengchuan Zhang, Qiuyuan Huang, Han Zhang, Zhe Gan, Xiaolei Huang, and Xiaodong He.
\newblock Attngan: Fine-grained text to image generation with attentional generative adversarial networks.
\newblock \emph{CVPR}, 2018.

\bibitem[Yang et~al.(2024)Yang, Yu, Meng, Xu, Ermon, and Bin]{yang2024mastering}
Ling Yang, Zhaochen Yu, Chenlin Meng, Minkai Xu, Stefano Ermon, and CUI Bin.
\newblock Mastering text-to-image diffusion: Recaptioning, planning, and generating with multimodal llms.
\newblock In \emph{Forty-first International Conference on Machine Learning}, 2024.

\bibitem[Zhang et~al.(2017)Zhang, Xu, Li, Zhang, Huang, Wang, and Metaxas]{zhang2017stackgan}
Han Zhang, Tao Xu, Hongsheng Li, Shaoting Zhang, Xiaolei Huang, Xiaogang Wang, and Dimitris Metaxas.
\newblock Stackgan: Text to photo-realistic image synthesis with stacked generative adversarial networks.
\newblock \emph{ICCV}, 2017.

\bibitem[Zhang et~al.(2021)Zhang, Koh, Baldridge, Lee, and Yang]{zhang2021cross}
Han Zhang, Jing~Yu Koh, Jason Baldridge, Honglak Lee, and Yinfei Yang.
\newblock Cross-modal contrastive learning for text-to-image generation.
\newblock In \emph{CVPR}, 2021.

\bibitem[Zhang et~al.(2023)Zhang, Yang, Feng, Qin, Chen, Yu, Chen, Wang, Savarese, Ermon, et~al.]{zhang2023hive}
Shu Zhang, Xinyi Yang, Yihao Feng, Can Qin, Chia-Chih Chen, Ning Yu, Zeyuan Chen, Huan Wang, Silvio Savarese, Stefano Ermon, et~al.
\newblock Hive: Harnessing human feedback for instructional visual editing.
\newblock \emph{arXiv preprint arXiv:2303.09618}, 2023.

\bibitem[Zhang et~al.(2024)Zhang, Yang, Cai, Yu, Xie, Tian, Xu, Tang, Yang, and Cui]{zhang2024realcompo}
Xinchen Zhang, Ling Yang, Yaqi Cai, Zhaochen Yu, Jiake Xie, Ye~Tian, Minkai Xu, Yong Tang, Yujiu Yang, and Bin Cui.
\newblock Realcompo: Dynamic equilibrium between realism and compositionality improves text-to-image diffusion models.
\newblock \emph{arXiv preprint arXiv:2402.12908}, 2024.

\bibitem[Zhou(2004)]{zhou2004image}
Wang Zhou.
\newblock Image quality assessment: from error measurement to structural similarity.
\newblock \emph{IEEE transactions on image processing}, 13:\penalty0 600--613, 2004.

\bibitem[Zhu et~al.(2019)Zhu, Pan, Chen, and Yang]{zhu2019dm}
Minfeng Zhu, Pingbo Pan, Wei Chen, and Yi~Yang.
\newblock Dm-gan: Dynamic memory generative adversarial networks for text-to-image synthesis.
\newblock In \emph{CVPR}, 2019.

\end{thebibliography}

\end{document}